%% file: neurips_2026.tex
\documentclass{article}

\PassOptionsToPackage{numbers, compress}{natbib}

\usepackage[eandd, preprint]{neurips_2026}

\usepackage[utf8]{inputenc} 
\usepackage[T1]{fontenc}    
\usepackage{hyperref}       
\usepackage{url}            
\usepackage{booktabs}       
\usepackage{amsfonts}       
\usepackage{nicefrac}       
\usepackage{microtype}      
\usepackage[table]{xcolor}  
\usepackage{multirow}
\usepackage{amsmath}
\usepackage{graphicx}
\usepackage{subcaption}
\usepackage{float}
\usepackage{tikz}
\usepackage{tabularx}
\usepackage{enumerate}
\usepackage{enumitem}
\usepackage{array}
\usepackage{etoolbox}
\usepackage{pifont} 
\usepackage{wrapfig}
\usepackage{makecell}

\usepackage{pgfplots}
\usepgfplotslibrary{groupplots}
\pgfplotsset{
  compat=1.18,
  every axis/.append style={
    axis line style={draw=black!25, line width=0.35pt},
    legend style={draw=none},
    grid style={draw=black!8, line width=0.3pt},
  },
}

 \newcommand{\cmark}{\textcolor[rgb]{0.0,0.6,0.0}{\ding{51}}}
\newcommand{\xmark}{\textcolor{red}{\ding{55}}}

\usetikzlibrary{arrows.meta,positioning,decorations.pathreplacing}
\usepackage{placeins}
\usepackage[most]{tcolorbox}
\tcbset{
  colback=gray!5,
  colframe=gray!70,
  boxrule=0.5pt,
  arc=6.5pt,
  left=6pt,right=6pt,top=4pt,bottom=4pt,
}
\newtcolorbox{highlightbox}{}

\title{SituRecBench: Evaluating Frontier Models as Situated Recommendation Assistants in 3D Environments}

\author{
  Jingtong Yang$^{1}$\thanks{Equal contribution.}
  \quad
  Dongding Lin$^{1}$\footnotemark[1]
  \quad
  Jian Wang$^{2}$\thanks{Corresponding author.}
  \quad
  Wenjie Li$^{1}$ \\
  $^{1}$Department of Computing, The Hong Kong Polytechnic University \\
  $^{2}$College of Computer Science, Sichuan University \\
  \texttt{jyang9.yang@connect.polyu.hk}
  \quad
  \texttt{dongding88.lin@connect.polyu.hk} \\
  \texttt{wangjian51@scu.edu.cn}
  \quad
  \texttt{cswjli@comp.polyu.edu.hk} \\ \\
  Project Page: \url{https://siturec.github.io/SituRecBench/}
}

\begin{document}

\maketitle

\begin{abstract}

Recommender systems are transitioning from static retrieval engines into interactive assistants grounded in physical environments. However, current studies largely neglect the combined complexities of spatial navigation, visual grounding, and multi-turn social interaction. 
To bridge this gap, we introduce \textbf{SituRecBench}\footnote{Code, data, and demos are all available at \url{https://siturec.github.io/SituRecBench/}.}, an interactive benchmark for evaluating frontier vision-language models (VLMs) as situated recommendation assistants within high-fidelity 3D environments. 
We reformulate recommendation as a situated assistance process in which an assistant interacts with a user while sharing the user's egocentric view.
Our benchmark simulates real-world shopping environments (e.g., supermarkets and home furnishings stores), requiring assistants to elicit preferences, ground recommendations in the environment, and guide the user to relevant locations through dialogue. 
The benchmark evaluates dialogue quality, recommendation utility, and navigation effectiveness through a unified interactive protocol.
Extensive experiments reveal persistent challenges in spatial grounding and preference elicitation, highlighting limitations in their ability to integrate global contextual information with local observations.
SituRecBench provides a rigorous foundation for advancing the development of situated assistants in the real world.

\end{abstract}

\input{sections/1_introduction}
\input{sections/2_formulation}

\input{sections/3_benchmark}
\input{sections/4_experiments}

\input{sections/5_related_work}
\input{sections/6_limitations}
\input{sections/7_conclusion}


\newpage
\section*{Acknowledgements}
This work was supported by the General Research Fund (GRF) of the Research Grants Council of Hong Kong (PolyU 15207122 and PolyU 15205325), Fundamental Research Funds for the Central Universities (YJ202620), and also in part by the PolyU Postdoc Matching Fund (4-W40Z). The authors would like to sincerely thank Ruiman Huang from The Hong Kong Polytechnic University for the development of the human evaluation web tool, and thank the anonymous reviewers for their valuable feedback and constructive suggestions.

\bibliography{reference}
\bibliographystyle{unsrt}


\newpage
\appendix
\input{sections/appendix}


\end{document}

%% file: sections/1_introduction.tex
\section{Introduction}

Conversational recommendations are undergoing a paradigm shift from presenting items statically to acting as interactive assistants~\cite{lin2024screen,zhang2025bootstrapping,lin2026and}. 
In real-world scenarios, such as a shopping assistant helping a user in a retail store, recommendation is an inherently physical and dynamic process. 
A useful assistant must not only retrieve relevant products but also ask clarifying questions, infer user preferences, help the user navigate the environment, and guide the user toward appropriate items. 
The user, in turn, provides feedback through both dialogue and physical actions. 
As large language models (LLMs) have emerged as powerful foundations for intelligent agents, developing LLM-based situated recommendation assistants underscores a promising and practical research direction.

\begin{figure}[th!]
\centering
    \includegraphics[width=\textwidth]{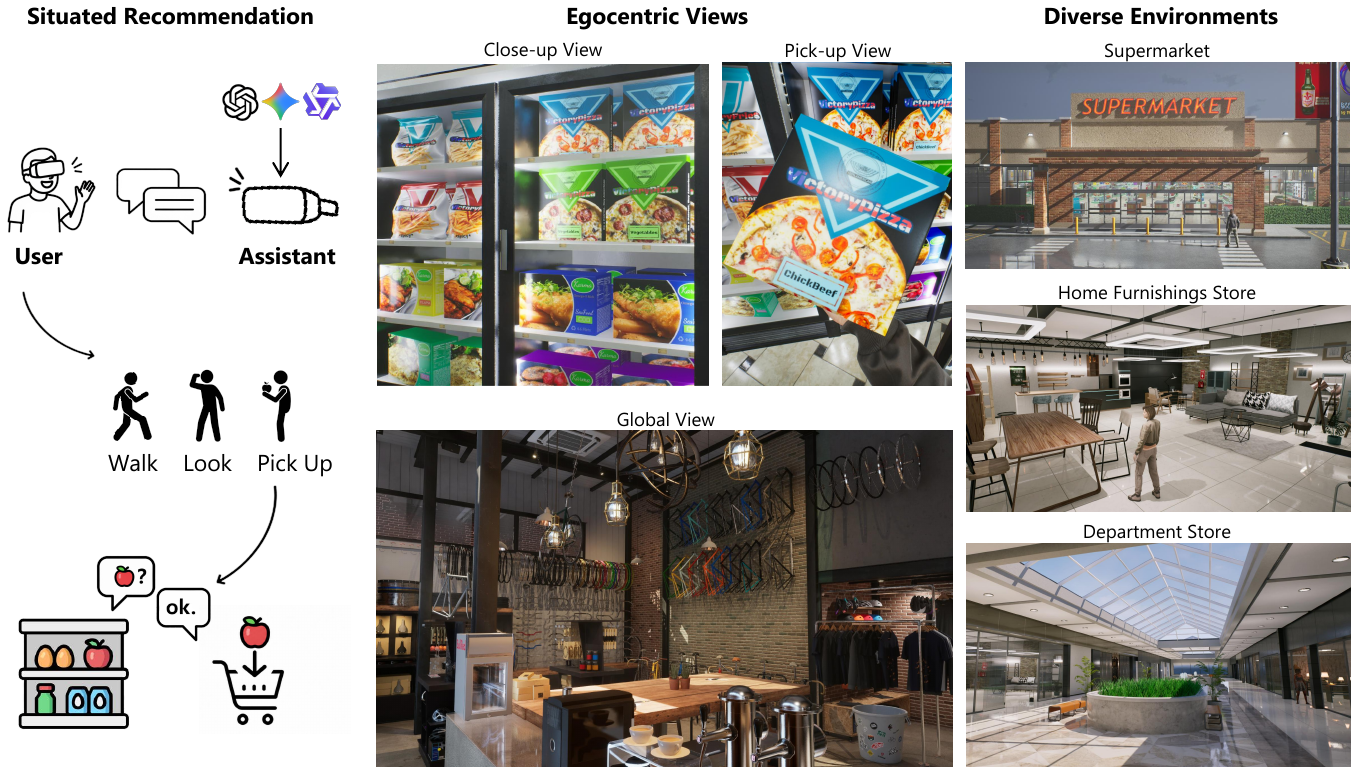}
    \caption{Our SituRecBench aims to (1) provide rich 3D interactive environments (e.g., supermarkets, home furnishings stores); (2) share the same egocentric view with the user and provide sufficient support for VLM agents as recommendation assistants; and (3) facilitate the evaluation and development of situated agents.}
    \label{fig:intro-overview}
    \vspace{-6pt}
\end{figure}

Despite its great potential, existing benchmarks provide limited support for evaluating these capabilities jointly (see Table~\ref{tab:benchmark-comparison}).
Standard situated conversation benchmarks~\cite{moon2020situated,kottur2023overview,lin2024screen,wang2025muse} focus heavily on dialogue and preference elicitation, but abstract away visual grounding and physical constraints. 
Conversely, physical interaction benchmarks provide rich embodied environments for navigation and manipulation~\cite{du2024constrained,hu2025marketgen,ye2025simworld}, but their objectives are typically defined by explicit instructions rather than implicit user preferences and recommendation utility. 
Overall, existing benchmarks are insufficient for jointly evaluating spatial navigation, visual grounding, and multi-turn interaction in situated recommendation.

To meet the growing demands, we introduce \textbf{SituRecBench}, a novel benchmark for developing and evaluating situated recommendation. Built on Unreal Engine 5, SituRecBench drops assistants into simulated 3D shopping environments, including supermarkets and home furnishings stores populated with items and structured metadata. 
Leveraging powerful large language models (LLMs) and vision-language models (VLMs)~\cite{zhang2024vision}, we set up a simulated user with specific profiles and goals.
As shown in Figure~\ref{fig:intro-overview}, the user simulator provides both language expressions and physical actions, such as walking and picking up items. 
The recommendation assistant and the user share the same egocentric view, simulating real-world assistants enabled by wearable devices such as smart glasses.
Operating over global contextual information (e.g., environment spaces, item attributes, initial instructions) and an egocentric view, the assistant must elicit user needs, ground recommendations in
the environment, and guide the user to relevant locations through dialogue.

Based on SituRecBench, we aim to conduct a comprehensive assessment of state-of-the-art VLMs as recommendation assistants~\cite{openai2024gpt4o,openai2026gpt56,gemini2026gemini31,qwen36plus,anthropic2026opus5}. 
Existing evaluation protocols in conversational recommendation are limited to static judgment~\citep{bernard2025limitations}, without reflecting the dynamic, interactive nature of situated recommendation.
An assistant might recommend a highly relevant item but exhaust the user's patience through redundant questioning, or navigate efficiently but violate the user's budget. 
To this end, our protocol assesses these crucial dimensions: overall outcome, dialogue quality, recommendation utility, and navigation effectiveness. 
Our experiments reveal that frontier VLMs still struggle significantly in situated recommendation settings, frequently exhibiting poor spatial grounding, premature recommendations, and ineffectiveness in recovering from user rejections. 
Through ablation studies and error analysis, we find that this may largely stem from the models' limited ability to integrate global context with local observations.

In summary, our contributions are as follows:
1) We instantiate the task of \textbf{situated recommendation}, where the recommendation assistant and user share the same egocentric view in a 3D environment. The assistant must infer user preferences and assist users through multi-turn conversational and physical interactions.
2) We introduce SituRecBench, a novel benchmark with \textbf{simulated, interactive 3D environments} containing egocentric views, structured item metadata, user profiles, shopping goals, and physical actions. 
3) We propose a unified \textbf{evaluation protocol} that employs both user simulators and human evaluators to assess state-of-the-art VLMs, highlighting critical bottlenecks in current models when deployed as situated assistants. 
Our work provides valuable insights into serving as a practical proxy for developing real-world situated agents in the future.

%% file: sections/2_formulation.tex
\section{Problem Definition}
\label{sec:definition}

\begin{table}[t!]
\centering
\caption{Comparison of SituRecBench with related benchmarks. \cmark~and \xmark~indicate the presence and absence of a feature, respectively. ``Env.'', ``Ego.'', ``Phys.'', ``Rec.'', ``Eval.'' are short for environment, egocentric, physical, recommendation, and evaluation, respectively. See Appendix~\ref{app:comparison-criteria} for definitions of the comparison criteria.}
\label{tab:benchmark-comparison}
\resizebox{1.0\textwidth}{!}{
\begin{tabular}{l c c c c c c c}
\toprule
\textbf{Benchmark} & \makecell{\textbf{Spatial} \\ \textbf{Env.}} & \makecell{\textbf{Ego.} \\ \textbf{View}}  & \makecell{\textbf{Dialog}} & \makecell{\textbf{Phys.} \\ \textbf{Action}} & \makecell{\textbf{Rec.}} & \makecell{\textbf{Interactive} \\ \textbf{Eval.}} & \makecell{\textbf{Scenario} } \\
\midrule
SIMMC-VR~\cite{wu2023simmc} & \cmark & \cmark & \cmark & \xmark & \cmark & \xmark & VR Shopping \\
SCREEN~\cite{lin2024screen} & \cmark & \cmark  & \cmark & \xmark & \cmark & \xmark & Multimodal Rec.\\
MUSE~\cite{wang2025muse} & \xmark & \xmark & \cmark & \xmark & \cmark & \xmark & Multimodal Rec. \\
MarketGen~\cite{hu2025marketgen} & \cmark & \cmark & \xmark & \cmark & \xmark & \xmark & Embodied Retail \\
DeliveryBench~\cite{mao2026deliverybench} & \cmark & \cmark  &  \xmark   & \cmark   & \xmark  & \xmark   & Food Delivery  \\
S$^3$IT~\cite{sun2025s} & \cmark & \cmark & \cmark & \xmark & \xmark & \cmark & Situated Social Test \\
Alexa Arena~\cite{gao2023alexa} & \cmark & \cmark & \cmark & \cmark & \xmark & \cmark & Gamified Robotics \\
SimWorld~\cite{ye2025simworld} & \cmark & \cmark & \cmark & \cmark & \xmark & \xmark & Urban Robotics \\
Virtual Community~\cite{zhou2026virtual} & \cmark & \cmark & \cmark   & \cmark  & \xmark & \xmark & Embodied Community \\
\midrule
\textbf{SituRecBench (Ours)} & \cmark & \cmark & \cmark & \cmark & \cmark & \cmark & Situated Shopping \\
\bottomrule
\end{tabular}}
\end{table}

We study \emph{situated recommendation}, a novel setting in which an assistant must help a user satisfy an underspecified shopping need through multi-turn communication and physically grounded interaction in a 3D environment. 
Unlike conventional recommender systems that rank items from a static catalog,  the assistant and user share the same egocentric view, and the assistant must infer latent preferences from language, visual context, and user behavior before giving grounded recommendations (as illustrated in Figure~\ref{fig:intro-overview}).

\paragraph{Situated Recommendation as Interactive Assistance.}

We formalize situated recommendation as a partially observable interactive decision-making problem between a user and an assistant in a co-located physical environment. An episode is defined as:
\[
\mathcal{T} = \langle \mathcal{E}, \mathcal{I}, \mathcal{U}, \mathcal{A}, \mathcal{O}, \mathcal{S}, P, R, \Gamma \rangle,
\]
where $\mathcal{E}$ is a 3D environment, $\mathcal{I}$ is the item set, $\mathcal{U}$ is the user space, $\mathcal{A}$ is the assistant action space, $\mathcal{O}$ is the observation space, and $\mathcal{S}$ is the environment state space. The transition function $P$, utility $R$, and termination rule $\Gamma$ define the interaction dynamics and evaluation.

The environment includes scene layout, traversable space, and contextual factors; each item $i \in \mathcal{I}$ is associated with category, attributes, price, location, visual appearance, and affordances. At the beginning of an episode, we use a user configuration $u = \langle z_u, g_u, b_u, \rho_u \rangle$, where $z_u$ denotes latent preferences, $g_u$ the shopping goal, $b_u$ the budget, and $\rho_u$ behavioral traits. The assistant does not directly observe $z_u$, $g_u$, or $b_u$, and must infer them online.

\paragraph{State, Observation, and History.}

At each turn $t$, the state is defined as:
\[
s_t = \langle q_t, \theta_t, \mathcal{B}_t, \mathcal{K}_t, \mathcal{I}_t, h_t \rangle \in \mathcal{S},
\]
where $q_t$ and $\theta_t$ denote the user's pose, $\mathcal{B}_t$ the current basket, $\mathcal{K}_t$ the revealed item information, $\mathcal{I}_t$ the current object configuration, and $h_t$ the interaction history. The assistant receives a partial observation $o_t = \Omega(s_t)$, consisting of egocentric visual input, dialogue history, visible objects, and optionally structured metadata, depending on the evaluation setting. The history records previous observations, assistant actions, and resulting user/environment feedback, such as clarifications, rejections, movements, inspections, and purchases.

\begin{figure}[!t]
\centering
    \includegraphics[width=\textwidth]{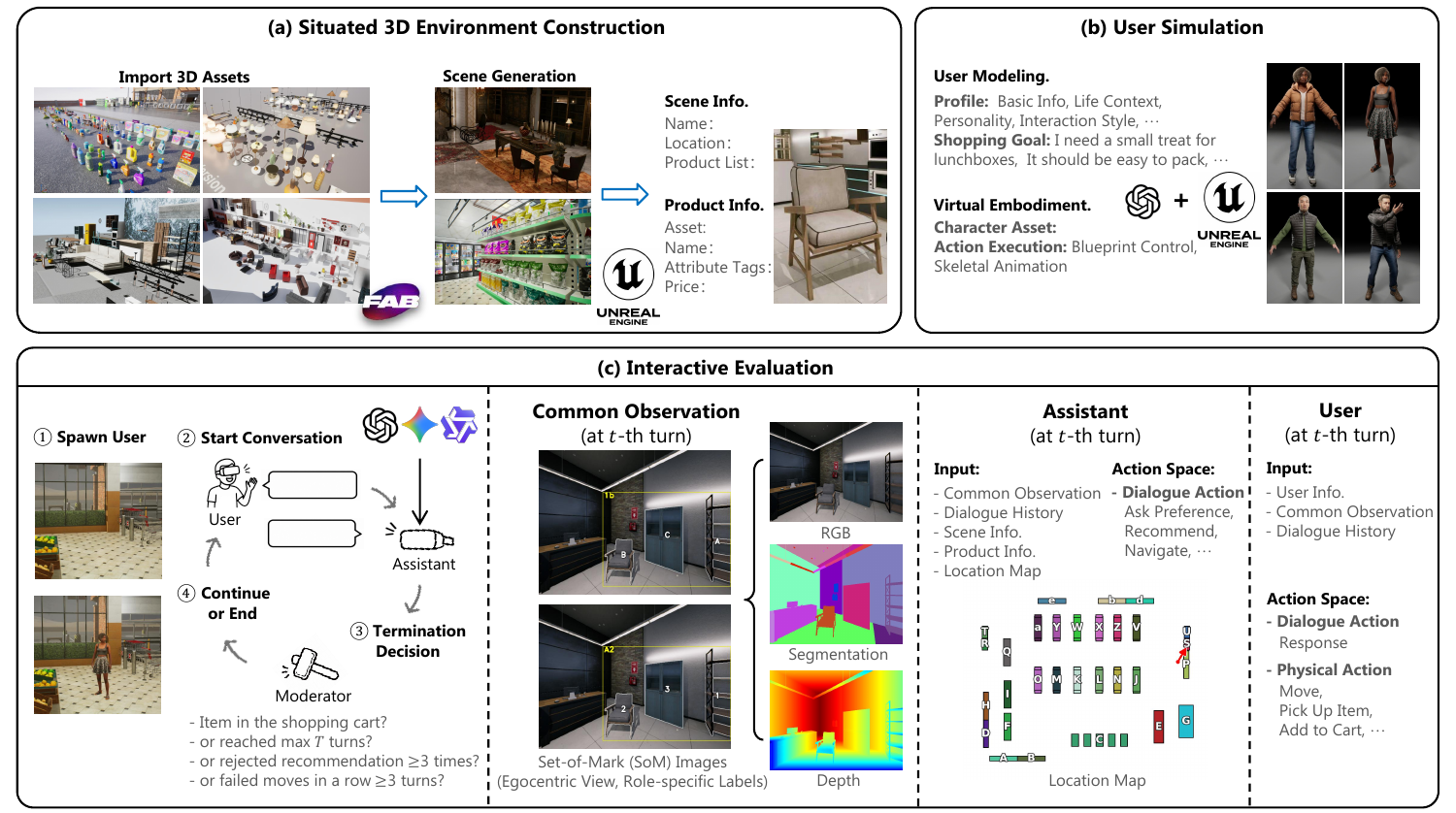}
    \caption{Overview of SituRecBench. (a) Situated 3D shopping environment construction; (b) User simulation with profiles, goals, and physical actions; and (c) Interactive evaluation of VLM-based recommendation assistants.}
    \label{fig:benchmark-overview}
    \vspace{-6pt}
\end{figure}

\paragraph{Assistant Action Space.}

A situated recommendation assistant is a policy given by:
\[
\pi_\theta(a_t \mid h_t, o_t),
\]
which maps the current observation and interaction history to an assistant action. Unlike static recommendations, the assistant action space is hybrid:
\[
a_t = \langle a_t^{\mathrm{lang}}, a_t^{\mathrm{rec}}, a_t^{\mathrm{nav}}, a_t^{\mathrm{int}} \rangle.
\]
Here, $a_t^{\mathrm{lang}}$ covers communicative actions such as asking or explaining, $a_t^{\mathrm{rec}}$ covers item- or plan-level recommendation, $a_t^{\mathrm{nav}}$ covers navigation guidance, and $a_t^{\mathrm{int}}$ covers interaction instructions such as inspect, pick up, or add to cart. This reflects the core challenge of situated recommendation: decide not only \emph{what} to recommend, but also \emph{when} to ask, \emph{where} to guide the user, \emph{how} to communicate the recommendation, and \emph{which} physical interaction should happen next.

\paragraph{User Action and Environment Transition.}

Given the user's current observation $o_t^{\mathrm{user}}$, interaction history $h_t$, and the assistant's utterance $a_t^{\mathrm{lang}}$, the user simulator produces a response
\[
y_t \sim \pi_u(y_t \mid o_t^{\mathrm{user}}, h_t, a_t^{\mathrm{lang}}, u),
\]
where $y_t$ may include a natural-language reply, an acceptance or rejection decision, a physical action, or an updated intent. The environment then transitions according to
\[
s_{t+1} \sim P(s_{t+1} \mid s_t, a_t, y_t).
\]

\paragraph{Objective.}

The goal of situated recommendation is to find an assistant policy $\pi_\theta$ that maximizes the expected episode-level utility:
\[
\max_{\pi_\theta} \; \mathbb{E}_{\mathcal{T}, u, \tau \sim (\pi_\theta, \pi_u, P)}
\left[
R(\tau; u, \mathcal{E}, \mathcal{I})
\right],
\]
where $\tau$ is the full interaction trajectory. The utility can be decomposed as an evaluation function on overall outcome, dialogue quality, recommendation utility, and navigation effectiveness. 

An episode terminates when one of the conditions in $\Gamma$ is met: $\Gamma(s_t, h_t) = 1$.
Termination conditions include a cart addition, moderator termination, motion stalls, or reaching the turn or recommendation limit. This makes situated recommendation a closed-loop evaluation problem: the assistant's earlier recommendations shape the user's later observations, actions, decisions, and final satisfaction.

%% file: sections/3_benchmark.tex
\section{SituRecBench}
\label{sec:benchmark}

SituRecBench supports shopping interactions between a recommendation assistant and a simulated user in two retail environments, enabling interactive evaluation of the assistant (Figure~\ref{fig:benchmark-overview}).

\subsection{Situated 3D Environment Construction}
\label{sec:env_construct}
We construct each 3D shopping environment and annotate product attributes and locations to support navigation, product interaction, and recommendation.

\paragraph{Environment Construction with Unreal Engine.}
We construct two retail environments with Unreal Engine (UE) 5~\citep{ue5}: \textit{Supermarket} and \textit{Home Furnishings Store}. Product placement in the Supermarket follows the layout of a typical supermarket, while the Home Furnishings Store organizes furniture by style and use into themed stores on a single floor.

\paragraph{Semantic Metadata Annotation.}
To provide the product information needed for recommendation, we match 3D product models to real-world retail listings and assign product names, attribute tags, and prices. We then link catalog entries to the corresponding scene objects, display fixtures, and spatial locations. Product sources, scene layouts, and annotation details are provided in Appendix~\ref{app:semantic-metadata}.

\subsection{User Simulation}
\label{sec:user_simulation}

\label{sec:user-profile-task-config}
\paragraph{User Modeling.}
To simulate how real users naturally communicate and behave during shopping, we prompt a vision-language model (VLM) to role-play a shopper with a user profile and a shopping goal. This design is motivated by prior findings that users' backgrounds, personalities, and preferences can guide the generation of personalized dialogue and behavior~\citep{wang2023target,wang2025user}, and that concrete needs and constraints give simulated users clear interaction objectives~\citep{yao2024tau}.

In our implementation, user profiles combine basic information and life context to establish each shopper's personal circumstances, with personality and interaction style shaping how they engage with the assistant. We complement these profiles with shopping goals that specify users' needs, preferences, and budgets for individual tasks. The design of these goals draws on consumer behavior research highlighting the influence of shopping purposes and situational factors on purchase behavior~\citep{belk1975situational}. We therefore use shopping themes to represent different purposes and everyday contexts, and combine these themes with the products available in each scene to formulate shopping goals. Each profile is paired with multiple tasks, allowing the same shopper to pursue different goals. Task construction, field definitions, and behavioral rules are detailed in Appendix~\ref{app:user-profiles}.

\paragraph{Virtual Embodiment.}
In real-world shopping, users navigate the store, look around, and interact with products as they gather information and make shopping decisions. We simulate these behaviors with UE5 character assets, using Blueprints to control basic actions and skeletal animations to represent the corresponding body movements and poses. At runtime, UnrealCV invokes the Blueprint functions to execute these actions, while the engine's movement and collision systems ensure that users navigate within the store layout while avoiding obstacles. As users move and interact, the available visual information changes, informing their subsequent actions and shopping decisions. Implementation details are provided in Appendices~\ref{app:user-embodiment}--\ref{app:action-space}.

\subsection{Observation and Action Spaces}
\label{sec:observation_action}

The user and assistant share the current first-person view and interaction history but receive different additional information. The user has access to their private profile and shopping goal, while the assistant receives catalog, map, and location information to guide shopping.

At turn \(t\), their shared egocentric observation is:
\[
\mathcal{O}_t^{\mathrm{ego}}
= \{\mathcal{O}_t^{\mathrm{rgb}}, \mathcal{H}_t\},
\]
where \(\mathcal{O}_t^{\mathrm{rgb}}\) is the first-person RGB view and \(\mathcal{H}_t\) contains dialogue and previous user actions. We label visible products and spatial targets in each role's first-person view using Set-of-Mark (SoM)~\citep{yang2023set} overlays to support interaction and navigation. The full observations are
\[
\mathcal{O}_t^{u}
= \{\mathcal{O}_t^{\mathrm{som},u}, \mathcal{H}_t, \mathcal{M}_t^{\mathrm{user}},
\mathcal{P}^{\mathrm{profile}}, \mathcal{G}^{\mathrm{goal}}\},
\]
and
\[
\mathcal{O}_t^{s}
= \{\mathcal{O}_t^{\mathrm{som},s}, \mathcal{H}_t, \mathcal{M}_t^{s},
L_t^{u}, I^{\mathrm{product}}, I^{\mathrm{loc}}\}.
\]
Here, \(\mathcal{M}_t^r\) maps markers to entities for role \(r\); \(\mathcal{P}^{\mathrm{profile}}\) and \(\mathcal{G}^{\mathrm{goal}}\) are the user's private profile and simulator-facing goal fields. The assistant receives the layout map with the user's pose \(L_t^{u}\), the full product catalog \(I^{\mathrm{product}}\), and the location index \(I^{\mathrm{loc}}\). The location index links products to fixtures in the Supermarket and to stores and display areas in the Home Furnishings Store.

The SoM overlays mark products with contours and buildings with bounding boxes, together with corresponding labels, on the first-person RGB view, using depth to limit the annotation distance. These labels allow models to specify action targets without predicting coordinates. To avoid label-sharing shortcuts, the two roles use different label mappings and cannot disclose labels in dialogue. Rendering and numbering details are in Appendix~\ref{app:observation-interface}.

\paragraph{Assistant Actions.}
The assistant primarily performs \textit{dialogue acts}, such as greeting the user, asking about preferences, clarifying needs, guiding navigation, recommending and explaining products, answering questions, comparing options, responding to rejection or hesitation, and confirming product suitability or purchase intent.

\paragraph{User Actions.}
User actions include both \textit{dialogue} and \textit{physical} actions.
At each turn, the user responds to the assistant in natural language and selects one physical action. The environment executes the action and updates the scene state and the user's view. Action definitions and execution details are provided in Appendices~\ref{app:user-embodiment}--\ref{app:action-space}.

Table~\ref{tab:bench_statistics} presents the statistical overview of the curated SituRecBench environments. Specifically, ``Products'' denotes unique catalog items, ``Instances'' refers to 3D scene objects registered in the environment, whereas ``Nav. Targets'' denote unique navigation locations. Detailed counting conventions and metadata definitions are provided in Appendix~\ref{app:semantic-metadata}.

\subsection{Evaluation Protocol}
\label{sec:evaluation}

Building on the environments and user simulation described above, we establish an interactive evaluation protocol to assess how effectively assistants support users throughout shopping.

\begin{wraptable}{r}{0.55\textwidth}
\centering
\captionof{table}{Statistics of SituRecBench.}
\label{tab:bench_statistics}
{\footnotesize
\setlength{\tabcolsep}{3pt}
\begin{tabular*}{\linewidth}{@{\extracolsep{\fill}}lccc@{}}
\toprule
\textbf{Environment} & \textbf{Supermarket} & \makecell{\textbf{Home Furnishings} \\ \textbf{Store} } & \textbf{Total} \\
\midrule
\# Products & 233 & 222 & 455 \\
\# Instances & 1,324 & 244 & 1,568 \\
\# Nav. Targets & 29 & 31 & 60 \\
\# Waypoints & 42 & 38 & 80 \\
\# User Profiles & 25 & 25 & 50 \\
\# Tasks & 100 & 100 & 200 \\
\bottomrule
\end{tabular*}}
\end{wraptable}

\paragraph{Interaction Loop.}
Each episode starts with a profile--goal pair and the user at the scene's starting point. ChatArena~\citep{ChatArena} coordinates assistant--user turns through the observation and action spaces defined in Section~\ref{sec:observation_action}. The Python runtime executes user actions in UE5 and retrieves subsequent observations through UnrealCV~\citep{DBLP:conf/mm/QiuZZQXKW17}, allowing the assistant to respond to the updated situation. Episodes end upon cart addition, moderator termination, motion stalls, or turn/recommendation limits. Further implementation details are provided in Appendices~\ref{app:unrealcv-runtime} and~\ref{app:interaction-protocol}.

\paragraph{Evaluation Metrics.}
Using interaction trajectories, we evaluate assistant performance in four dimensions: overall outcome, dialogue quality, recommendation utility, and navigation effectiveness. For overall outcome, we report Success Rate (\textbf{SR}), the proportion of interactions ending with an item added to the cart, and the mean number of recorded turns per interaction (\textbf{Turn}).

Dialogue quality is assessed by an LLM judge, which rates each complete interaction based on its textual record~\citep{wang2023target,wang2025muse}. The judge scores each of seven dimensions from 1 to 5: Informativeness (\textbf{Inform.}), Naturalness (\textbf{Natur.}), Groundedness (\textbf{Ground.}), Preference Alignment (\textbf{Prefer.}), Proactiveness (\textbf{Proact.}), Non-repetition (\textbf{Nonrep.}), and Overall Satisfaction (\textbf{Overall}).

Following prior situated recommendation work~\citep{lin2024screen,lin2026re2a}, we measure recommendation utility through recall and ranking quality relative to the user's purchased products. Recall at 5 (\textbf{R@5}) measures the proportion of purchased products appearing among the top five recommendations. Normalized Discounted Cumulative Gain at 10 (\textbf{NDCG@10}) gives more credit when these products rank higher among the top ten recommendations.

For navigation, we examine whether users follow the assistant's guidance and whether its navigation targets become visible~\citep{wu2023simmc}. Action Following Rate (\textbf{AFR}) measures how often the user's action type matches the assistant's suggestion. Local Navigation Alignment (\textbf{LNSR}) further checks whether local parameters, such as the direction or destination, also match. Cumulative Goal Visibility (\textbf{CGV}) measures how often the assistant's navigation targets appear in the user's current or subsequent views. Detailed scoring criteria and computation procedures for all metrics are provided in Appendix~\ref{app:metric-definitions}.

\paragraph{Interactive Human Evaluation.}
To examine whether simulated evaluation reflects assistant performance with human users, we develop an interface for shopping tasks using the same assistant models, settings, and interaction protocol as automatic evaluation. Participants fill in a user profile and are assigned randomly sampled tasks. They interact with each assistant through dialogue and actions in the virtual environment. During this process, we randomize assistant order and conceal model identities to reduce order effects and bias from prior impressions of the models. After each interaction, participants complete a five-point questionnaire about their shopping experience. Using the same metrics and LLM judge for dialogue ratings, we compare assistant performance with human and simulated users to assess whether simulation preserves relative assistant performance. We also compare participants' questionnaire responses with the judge's ratings of the same human interaction trajectories to assess whether automated judgments reflect participants' experience. Appendix~\ref{app:human-validation} details the procedures; Section~\ref{sec:human_evaluation} reports the experimental scale and results.

%% file: sections/4_experiments.tex
\begin{table}[t!]
\centering
\setlength{\tabcolsep}{3pt}
\caption{Evaluation results of different VLM models as situated recommendation assistants, with GPT-5.6 Sol acting as the simulated user. Bold marks the best performance per environment.}
\label{tab:main_results}
{\small
\begin{tabular*}{\linewidth}{@{\extracolsep{\fill}}llcccccc@{}}
\toprule
\multirow{2}{*}{\textbf{Environment}} & \multirow{2}{*}{\textbf{Model}} & \multicolumn{2}{c}{\textbf{Outcome}} & \multicolumn{2}{c}{\textbf{Navigation}} & \multicolumn{2}{c}{\textbf{Recommendation}} \\
\cmidrule(lr){3-4}\cmidrule(lr){5-6}\cmidrule(lr){7-8}
 & & \textbf{SR (\%)} & \textbf{Turn} & \textbf{AFR (\%)} & \textbf{LNSR (\%)} & \textbf{R@5 (\%)} & \textbf{NDCG@10 (\%)} \\
\midrule
\multirow{7}{*}{Supermarket} & Gemma 4 E4B & 28.0$_{\pm 5.7}$ & 9.0$_{\pm 0.7}$ & 49.2$_{\pm 2.0}$ & 10.0$_{\pm 3.7}$ & 17.3$_{\pm 4.5}$ & 15.0$_{\pm 4.4}$ \\
 & Qwen3-VL-8B & 34.0$_{\pm 9.5}$ & 11.8$_{\pm 1.2}$ & 34.5$_{\pm 3.6}$ & 8.8$_{\pm 3.0}$ & 20.3$_{\pm 4.6}$ & 17.7$_{\pm 5.0}$ \\
 & GPT-4o & 45.0$_{\pm 6.8}$ & 12.5$_{\pm 0.6}$ & 62.3$_{\pm 2.1}$ & 15.1$_{\pm 3.2}$ & 31.6$_{\pm 2.3}$ & 26.6$_{\pm 1.5}$ \\
 & Qwen3.6-Plus & 65.0$_{\pm 11.9}$ & 12.1$_{\pm 1.7}$ & 65.9$_{\pm 4.3}$ & 28.5$_{\pm 5.6}$ & 49.6$_{\pm 10.2}$ & 42.4$_{\pm 9.1}$ \\
 & Claude Opus 5 & 66.0$_{\pm 10.1}$ & 11.6$_{\pm 1.1}$ & 78.6$_{\pm 5.5}$ & 41.0$_{\pm 5.0}$ & 49.8$_{\pm 10.7}$ & 43.5$_{\pm 9.4}$ \\
 & GPT-5.6 Sol & 69.0$_{\pm 20.8}$ & 10.9$_{\pm 1.5}$ & \textbf{89.0}$_{\pm 1.2}$ & \textbf{53.5}$_{\pm 3.9}$ & 56.3$_{\pm 16.0}$ & 48.2$_{\pm 13.2}$ \\
 & Gemini 3.1 Pro & \textbf{80.0}$_{\pm 5.7}$ & 9.8$_{\pm 0.8}$ & 80.1$_{\pm 1.8}$ & 48.4$_{\pm 6.4}$ & \textbf{64.4}$_{\pm 2.1}$ & \textbf{56.0}$_{\pm 1.1}$ \\
\midrule
\multirow{7}{*}{\makecell[l]{Home\\Furnishings\\Store}} & Qwen3-VL-8B & 29.0$_{\pm 8.9}$ & 8.3$_{\pm 1.1}$ & 47.0$_{\pm 6.1}$ & 26.3$_{\pm 8.6}$ & 15.3$_{\pm 5.4}$ & 13.7$_{\pm 5.3}$ \\
 & Gemma 4 E4B & 31.0$_{\pm 8.2}$ & 9.6$_{\pm 1.3}$ & 40.0$_{\pm 5.8}$ & 29.0$_{\pm 2.9}$ & 17.3$_{\pm 4.9}$ & 16.4$_{\pm 5.5}$ \\
 & GPT-4o & 35.0$_{\pm 10.0}$ & 9.9$_{\pm 1.7}$ & 56.9$_{\pm 4.1}$ & 53.0$_{\pm 5.8}$ & 23.0$_{\pm 8.7}$ & 20.2$_{\pm 6.9}$ \\
 & Claude Opus 5 & 50.0$_{\pm 9.5}$ & 12.0$_{\pm 0.8}$ & 71.4$_{\pm 2.2}$ & 68.1$_{\pm 3.9}$ & 37.0$_{\pm 6.0}$ & 33.8$_{\pm 6.3}$ \\
 & Qwen3.6-Plus & 54.0$_{\pm 9.5}$ & 11.6$_{\pm 1.1}$ & 71.6$_{\pm 4.8}$ & 70.9$_{\pm 3.9}$ & 39.4$_{\pm 5.3}$ & 35.5$_{\pm 4.7}$ \\
 & GPT-5.6 Sol & 65.0$_{\pm 8.2}$ & 11.4$_{\pm 1.1}$ & \textbf{82.2}$_{\pm 3.5}$ & \textbf{88.8}$_{\pm 2.4}$ & 48.1$_{\pm 6.5}$ & 43.0$_{\pm 6.2}$ \\
 & Gemini 3.1 Pro & \textbf{66.0}$_{\pm 7.7}$ & 9.5$_{\pm 0.3}$ & 81.8$_{\pm 1.9}$ & 83.2$_{\pm 4.1}$ & \textbf{49.8}$_{\pm 7.1}$ & \textbf{43.7}$_{\pm 7.5}$ \\
\bottomrule
\end{tabular*}
}
\end{table}

\section{Experiments}
\label{sec:experiments}

\subsection{Experimental Setup}
We evaluate several frontier vision-language models as assistants: Claude Opus 5~\cite{anthropic2026opus5}, Gemini 3.1 Pro~\cite{gemini2026gemini31}, GPT-4o~\cite{openai2024gpt4o}, GPT-5.6 Sol~\cite{openai2026gpt56}, Qwen3.6-Plus~\cite{qwen36plus}, Qwen3-VL-8B-Instruct (Qwen3-VL-8B)~\cite{bai2025qwen3}, and Gemma 4 E4B (IT) (Gemma 4 E4B)~\cite{team2026gemma}. 
In the main experiments, GPT-5.6 Sol serves as both user simulator and conversation judge. All assistants complete the same 100 tasks in each of the two environments, yielding 1,400 trajectories in total. 
Metric details and supplementary results with a GPT-4o user simulator appear in Appendices~\ref{app:metric-definitions} and~\ref{app:user-simulator-sensitivity}, respectively.

\subsection{Main Results}
\label{sec:results}

\begin{figure}[!t]
\centering
\includegraphics[width=0.96\linewidth]{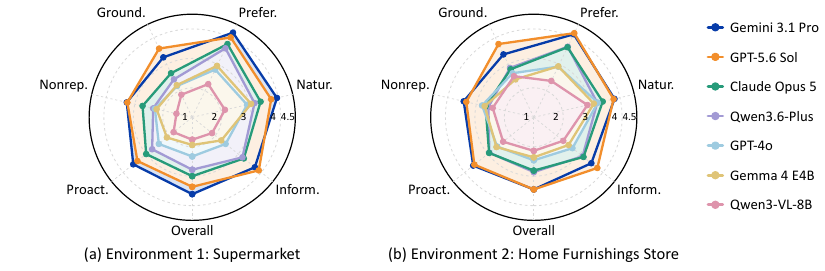}
\caption{Dialogue quality ratings of different VLMs as situated recommendation assistants across multiple evaluation dimensions, with GPT-5.6 Sol acting as the simulated user.}
\label{fig:cross_environment_radar}
\end{figure}

Table~\ref{tab:main_results} and Figure~\ref{fig:cross_environment_radar} summarize the assistants' task performance and dialogue quality across the two environments. GPT-5.6 Sol's responses are rated most informative, and its guidance is followed most consistently. Gemini 3.1 Pro, however, more often completes purchases and recommends items that match users' final choices in both environments. It also receives the highest overall satisfaction ratings in Supermarket. These contrasting strengths suggest that stronger action guidance does not always coincide with higher purchase success, highlighting the importance of assessing both how assistants guide users and whether they help complete a purchase. The two smaller open-weight models, Qwen3-VL-8B and Gemma 4 E4B, fall behind in both purchase completion and overall satisfaction, showing that their limitations affect the shopping experience as well as the outcome.

\begin{figure}[!t]
\centering
\includegraphics[width=0.99\linewidth]{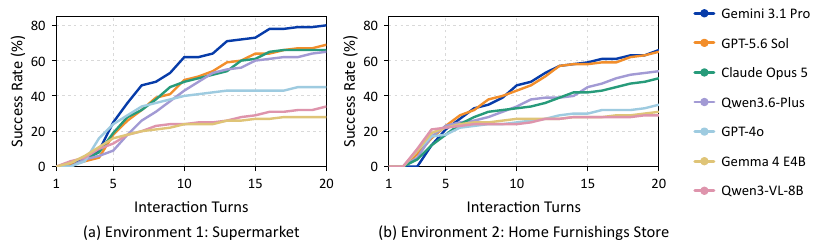}
\caption{Success rates of different assistants under varying interaction turns in two environments.}
\label{fig:cart_budget}
\end{figure}

Beyond these overall results, Figure~\ref{fig:cart_budget} shows how purchase success develops over successive interaction turns. In Home Furnishings Store, all models achieve similar purchase success within the first five turns. Gemini 3.1 Pro and GPT-5.6 Sol subsequently pull ahead, while Claude Opus 5 and Qwen3.6-Plus also continue to improve. Gains are smaller for GPT-4o and the two smaller models. Similarly, in Supermarket, Gemini establishes and maintains a lead midway through the interactions. The growing gaps suggest that much of the final performance difference emerges during continued interaction, when assistants need to keep moving the shopping process forward.

To complement these aggregate results, the qualitative cases in Appendix~\ref{app:qualitative-cases} illustrate how assistants respond to users' needs during individual interactions. GPT-4o acknowledges that a product exceeds the user's budget but still recommends it, leading to another rejection. In the furniture cases, one user accepts a chair for reading because it appears comfortable, while another rejects the same chair because its appearance does not suit the desired interior style. These examples highlight the need to align recommendations with users' constraints and intended product use.

\subsection{Interactive Human Evaluation Results}
\label{sec:human_evaluation}

Following the interactive human evaluation protocol in Section~\ref{sec:evaluation}, we compare 50 trajectories collected from five human participants on five randomly sampled Supermarket tasks with 500 GPT-5.6 Sol user trajectories from 100 task configurations, covering the same five assistants.

\begin{figure}[!t]
\centering
\includegraphics[width=\linewidth]{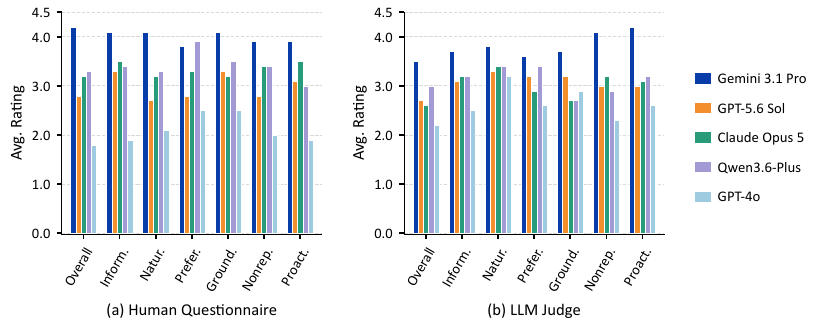}
\caption{Dialogue quality ratings of different assistants after interacting with real human evaluators in the Supermarket environment. (a) Using human questionnaire ratings; (b) Using the same LLM judge as automatic evaluation.}
\label{fig:human-questionnaire-judge-ratings}
\end{figure}

For the same human interactions, questionnaire ratings and automated scores yield similar model rankings across most dialogue dimensions, particularly informativeness, naturalness, and overall satisfaction (Figure~\ref{fig:human-questionnaire-judge-ratings}). This suggests that automated judging captures participants' relative assessments of assistants' dialogue quality reasonably well. Ranking agreement is lowest for groundedness: one possible reason is that human raters can check descriptions of product appearance and spatial location against the interaction images, whereas the LLM judge only reads textual records and cannot directly verify these visual details. This difference may lead to different assessments of groundedness.

Table~\ref{tab:human-validation-supermarket} also reports correlations in assistant performance between interactions with humans and simulated users. Human participants follow the suggested action types less often than simulated users. However, the two groups yield similar model rankings for overall satisfaction and local navigation alignment. In particular, Gemini 3.1 Pro receives the highest overall satisfaction rating in both groups. These results suggest that simulated users preserve assistants' relative performance on these dimensions reasonably well, supporting their use in interactive evaluation.

\begin{table}[!t]
\centering
\setlength{\tabcolsep}{1.5pt}
\caption{Correlations in assistant performance between interactions with humans and simulated users in the Supermarket environment.}
\label{tab:human-validation-supermarket}
{\small
\begin{tabular*}{\linewidth}{@{\extracolsep{\fill}}lcccccccccccc@{}}
\toprule
\textbf{Coefficient} & \makecell{\textbf{SR}} & \makecell{\textbf{AFR}} & \makecell{\textbf{LNSR}} & \makecell{\textbf{R@5}} & \makecell{\textbf{NDCG}\\\textbf{@10}}  & \makecell{\textbf{Overall}} & \makecell{\textbf{Inform.}} & \makecell{\textbf{Natur.}} & \makecell{\textbf{Ground.}} & \makecell{\textbf{Prefer.}} & \makecell{\textbf{Proact.}} & \makecell{\textbf{Nonrep.}} \\
\midrule
Pearson ($r$) & 0.800 & 0.287 & 0.742 & 0.619 & 0.574 & 0.773 & 0.774 & 0.726 & 0.698 & 0.807 & 0.753 & 0.744 \\
Spearman ($\rho$) & 0.462 & 0.100 & 0.800 & 0.300 & 0.300 & 0.700 & 0.359 & 0.667 & 0.564 & 0.700 & 0.600 & 0.900 \\
\bottomrule
\end{tabular*}}
\end{table}

\begin{table}[!t]
\centering
\caption{Effects of ablating model contexts on task performance and dialogue quality in the Supermarket environment.}
\label{tab:visual_ablation}
{\small
\begin{tabular*}{\linewidth}{@{\extracolsep{\fill}}lccccc@{}}
\toprule
\textbf{Condition} & \textbf{SR (\%)} & \textbf{AFR (\%)} & \textbf{LNSR (\%)} & \textbf{R@5 (\%)} & \textbf{NDCG@10 (\%)} \\
\midrule
Full Context & 69.0$_{\pm 20.8}$ & \textbf{89.0}$_{\pm 1.2}$ & 53.5$_{\pm 3.9}$ & \textbf{56.3}$_{\pm 16.0}$ & 48.2$_{\pm 13.2}$ \\
~~ \textit{w/o} Ego. Visual View (EVV) & \textbf{71.0}$_{\pm 11.9}$ & 84.8$_{\pm 3.5}$ & 42.4$_{\pm 10.2}$ & 55.5$_{\pm 13.2}$ & \textbf{49.5}$_{\pm 10.9}$ \\
~~ \textit{w/o} EVV \& Textual Caption & 65.0$_{\pm 6.8}$ & 69.9$_{\pm 3.1}$ & \textbf{58.7}$_{\pm 5.3}$ & 50.0$_{\pm 5.6}$ & 41.7$_{\pm 3.4}$ \\
\bottomrule
\end{tabular*}
}
\setlength{\tabcolsep}{1.5pt}
{\small
\begin{tabular*}{\linewidth}{@{\extracolsep{\fill}}lccccccc@{}}
\toprule
\textbf{Condition} & \textbf{Overall} & \textbf{Inform.} & \textbf{Natur.} & \textbf{Ground.} & \textbf{Prefer.} & \textbf{Proact.} & \textbf{Nonrep.} \\
\midrule
Full Context & \textbf{3.4}$_{\pm 0.4}$ & \textbf{3.9}$_{\pm 0.3}$ & \textbf{3.8}$_{\pm 0.3}$ & \textbf{3.6}$_{\pm 0.3}$ & \textbf{4.0}$_{\pm 0.3}$ & \textbf{3.4}$_{\pm 0.4}$ & \textbf{3.3}$_{\pm 0.4}$ \\
~~ \textit{w/o} EVV & \textbf{3.4}$_{\pm 0.3}$ & 3.7$_{\pm 0.2}$ & 3.7$_{\pm 0.2}$ & 3.5$_{\pm 0.3}$ & \textbf{4.0}$_{\pm 0.2}$ & \textbf{3.4}$_{\pm 0.4}$ & 3.1$_{\pm 0.3}$ \\
~~ \textit{w/o} EVV \& Textual Caption & 2.8$_{\pm 0.2}$ & 3.2$_{\pm 0.2}$ & 3.2$_{\pm 0.2}$ & 2.8$_{\pm 0.2}$ & 3.8$_{\pm 0.2}$ & 2.6$_{\pm 0.2}$ & 2.0$_{\pm 0.2}$ \\
\bottomrule
\end{tabular*}
}
\end{table}

\subsection{Ablation Study}
\label{sec:ablation_study}
To assess the contribution of first-person information, we evaluate three context settings with GPT-5.6 Sol as both assistant and user on the same 100 Supermarket tasks per condition, yielding 300 trajectories. With first-person image removed, the assistant achieves similar purchase and recommendation outcomes and receives comparable overall dialogue satisfaction ratings, but its navigation guidance is followed less closely. Further removing textual annotations lowers purchase and recommendation scores and all dialogue ratings, with the map and catalog still available. These results suggest that textual annotations provide the primary contribution of first-person information in this setting, helping the assistant connect the global catalog and map with visible products and navigation targets (Table~\ref{tab:visual_ablation}). Appendix~\ref{app:visual-ablation-full} details the ablation protocol.

%% file: sections/5_related_work.tex
\section{Related Work}

\paragraph{Conversational and Situated Recommendation.}
Situated recommendation builds on multimodal dialogue, where assistants use shared scene images to identify products mentioned by users and generate grounded responses~\cite{moon2020situated,kottur2023overview}. This setting has been extended to egocentric video, requiring assistants to track changes in the user's view and connect information across frames with the ongoing conversation~\cite{wu2023simmc}. To recommend suitable products, assistants also consider subjective preferences, situational factors, and user profiles alongside this visual context~\cite{long2023multimodal,lin2024screen,wang2025muse}. More recent approaches use dialogue and scene information to infer implicit and evolving preferences and select suitable scenes~\cite{lin2026and,lin2026re2a}. SituRecBench further extends this setting to product exploration and selection, evaluating how assistants support purchase decisions as users move through stores and inspect items.

\paragraph{Physical Environment Simulation.}
Existing studies create 3D scenes by arranging objects under semantic and spatial constraints~\cite{yang2024holodeck,feng2023layoutgpt,sun2025layoutvlm}. Within such environments, open-world platforms support physical and social activities among multiple agents~\cite{zhou2026virtual,ye2025simworld}. Dialogue-based simulation further incorporates communication, enabling agents to follow user instructions and visual assistants to guide users through physical tasks~\cite{gao2023alexa,zhang2025bootstrapping}. 
In retail settings, existing benchmarks evaluate navigation and fine-grained object manipulation, such as grasping and placing products~\cite{hu2025marketgen}. Similarly, SituRecBench works in interactive physical environments, with both dialogue and physical actions.

\paragraph{LLM-based User Simulation.}
LLM-based user simulation supports dialogue collection and assistant evaluation~\cite{ni2026survey}, using profiles and personalities to shape user responses~\cite{wang2023target,wang2025user}. Benchmarks use these interactions to evaluate assistants' understanding of user needs and preferences~\cite{qian2025userbench,kim2025cupid}, alongside task outcomes and user utility~\cite{wang2025training,bernard2025limitations}. To assess reliability, prior work compares relative assistant performance under simulated and human evaluations~\cite{dou-etal-2025-simulatorarena}. In comparison, SituRecBench leverages LLMs and physical engines to simulate dialogue and user behavior.

%% file: sections/6_limitations.tex
\section{Limitations and Potential Societal Impact}
\label{sec:limitation}

We acknowledge the limitations of this work as follows. SituRecBench relies on static egocentric views at each interaction turn, which limits its ability to capture dynamic visual changes during movement. 
Future work could integrate real-time video streams to better track these visual changes and understand user actions over time.
Furthermore, practical deployment demands low-latency responses and robust user privacy protection, both of which highlight the need for local deployment using lightweight VLMs. 
However, current lightweight VLMs exhibit limited performance on this benchmark, underscoring significant room for improvement before real-world adoption.

While SituRecBench focuses on retail recommendations, the underlying paradigm holds potential for broader applications in other situated assistance scenarios.
For instance, during museum visits, a situated assistant could dynamically adapt explanations and tour routes to individual interests. In hospital wayfinding, it could guide patients to target departments while navigating complex service procedures.
Future work could explore how assistants can effectively synthesize global environmental context with local egocentric observations to serve user needs across diverse real-world domains.

%% file: sections/7_conclusion.tex
\section{Conclusion}

In this paper, we introduce SituRecBench, a benchmark designed for situated recommendation assistance within shared egocentric 3D environments. We formalize this task as a partially observable interactive decision-making problem, accompanied by an evaluation framework spanning conversation, navigation, and recommendation.
Our experimental results reveal critical challenges in eliciting implicit preferences and proactively guiding user actions, particularly when assistants must interpret and adapt to dynamic behavioral feedback.
We hope SituRecBench serves as a valuable platform to foster future research toward building effective, real-world situated agents.

%% file: sections/appendix.tex
\providecommand{\promptslot}[1]{\texttt{\{\detokenize{#1}\}}}
\providecommand{\promptcode}[1]{\texttt{\detokenize{#1}}}

\section{Benchmark and Tasks}
\label{app:background-problem-definition}

\subsection{Benchmark Comparison Criteria}
\label{app:comparison-criteria}

Table~\ref{tab:benchmark-comparison} summarizes the tasks and evaluation features of the benchmarks compared. The definitions of the comparison criteria are as follows:
\begin{itemize} 

\item Spatial Environment (\textbf{Spatial Env.}): Whether the benchmark uses an environment with explicit spatial structure.

\item Egocentric View (\textbf{Ego. View.}): Whether the model receives egocentric (first-person) visual observations of the task environment as input.

\item \textbf{Dialog}: Whether multi-turn conversation with a user or social interlocutor forms part of the evaluated task.

\item Physical Action (\textbf{Phys. Action.}): Whether actions taken by the agent, or by a user following its guidance, are executed in the environment and change its state.

\item Recommendation (\textbf{Rec.}): Whether the benchmark evaluates recommending products or other items to meet a user's needs or preferences.

\item Interactive Evaluation (\textbf{Interactive Eval.}): Whether the evaluation includes online exchanges with a human or simulated user whose feedback informs the agent's subsequent responses or actions.

\item \textbf{Scenario}: The benchmark's main application or focused task setting.

\end{itemize}

\subsection{Formal Task Specification}
\label{app:formal-task-specification}

Building on the formulation in Section~\ref{sec:definition}, we specify the user configuration and assistant output used in the benchmark. A profile $p_u=\langle r_u,c_u,\phi_u,\rho_u\rangle$ contains the persona, demographic and living context, personality traits, and interaction parameters.

A shopping goal $g=\langle q_g,z_g,b_g,d_g,a_g\rangle$ contains the request, preferences, budget, disclosure policy, and acceptance policy. These fields are private inputs to the user simulator.

At each turn, the assistant implements the action components defined in Section~\ref{sec:definition} through a structured output:
\[
\tilde a_t=\langle C_t^{\mathrm{assistant}},
\mathbf D_t^{\mathrm{assistant}},
\widehat T_t^{\mathrm{user}},\mathcal R_t\rangle.
\]
Here, $C_t^{\mathrm{assistant}}$ is the user-facing response, $\mathbf D_t^{\mathrm{assistant}}$ records one primary dialogue action, $\widehat T_t^{\mathrm{user}}$ is the expected user action, and $\mathcal R_t$ is the internal Top-10 product ranking. The user receives the response text and independently selects the next action using the current observation, interaction history, and private configuration. The runtime executes the user's action and updates the environment state. Output fields and action constraints are detailed in Appendix~\ref{app:action-space}, and termination rules in Appendix~\ref{app:interaction-protocol}.

\FloatBarrier

\subsection{Environment and Task Construction}
\label{sec:appendix}

The scenes incorporate third-party building, product, and character assets, including assets obtained from UE Marketplace/Fab. Third-party assets are distributed solely as integrated components of the packaged Unreal Engine applications, in accordance with their respective licenses.

\paragraph{Environment Statistics and Scene Records.}
\label{app:benchmark-statistics}
\label{app:semantic-metadata}

Table~\ref{tab:bench_statistics} reports the benchmark statistics. The Supermarket has 233 products, 1,324 registered instances, 29 navigable targets, and 42 fixed waypoints; Home Furnishings Store has 222 products, 244 instances, 31 targets, and 38 waypoints. Each registered scene object counts as one instance, with multiple placements of the same product counted separately.

In the Supermarket, products are distributed across 29 fixtures according to storage requirements: 12 shelves for ambient foods and everyday goods, 3 chest freezers and 8 upright freezers for frozen foods, 4 open coolers for chilled foods and beverages, and 2 wooden produce stands for fresh fruit and vegetables.

Home Furnishings Store instead organizes products by style and use across seven stores: Atelier Prime (modern living-room furniture), Leather and Ink (study and office furniture), Light Logic (indoor lighting), Nordic Nest (Scandinavian-style everyday home furnishings), Play-cation (gaming, sports, and children's play equipment), The Plant (plants and planters), and Vicotian Curio Collection (classical furniture and vintage ornaments).

To connect these scene objects with recommendation and navigation, we maintain records for products and spatial entities. Product information comes primarily from Target and Walmart for Supermarket, and from Wayfair, The Home Depot, and IKEA for Home Furnishings Store. Each product record contains an internal product identifier, name, category, attribute tags, price, and segmentation color, and links to its scene instances and containing fixtures.

Records for stores, fixtures, and display areas contain their names, locations, spatial footprints, navigable entrances, and product lists. The layout map presents the spatial layout and displays the user's position and orientation during interaction.

\paragraph{User Profiles and Task Construction.}
\label{app:user-profiles}

Each scene has 25 user profiles, each paired with four shopping tasks, yielding 200 tasks across 50 profiles. Each profile contains a persona descriptor, demographic and living conditions, Big Five traits, and interaction settings: behavior style, patience, price sensitivity, explanation seeking, substitute willingness, navigation following, and talkativeness.

We construct shopping goals by combining shopping themes with the products available in each scene. For Supermarket, we draw inspiration from grocery shopping topic research~\citep{vega2022posterior} and develop themes such as breakfast preparation, household replenishment, dinner preparation, social gatherings, pet care, household cleaning, and holiday shopping. For Home Furnishings Store, we adapt persona archetypes from home-product research~\citep{yin2026copersona} to the available products, developing themes such as reading and nighttime lighting, home-office setup, storage, plant displays, family activities, and room furnishing.

We construct shopping tasks in four steps:
\begin{enumerate}
    \item \textbf{Define Roles and Shopping Themes.} Based on shopping contexts and the products available in each scene, we define role names, backgrounds, shopping themes, and shopping priorities. We use random seeds to sample gender, Big Five traits, and behavior style, and sample age group, employment, occupation, household, housing, transport, and storage conditions within the role's background constraints.
    \item \textbf{Select Target Products.} Based on each role's background and shopping theme, we select four target products from the catalog, considering coverage of categories, uses, and locations. Each target forms a separate task.
    \item \textbf{Specify Needs and Interaction Settings.} We define shopping needs and preferences around each target and select corresponding attribute labels from the product catalog. We then use random seeds to sample dialogue habits and a budget tier from options consistent with the role's behavioral characteristics.
    \item \textbf{Calculate Budgets.} We filter the catalog for products that meet the task's label requirements. The strict tier uses the target price as its starting point. For the moderate and more flexible tiers, we sort eligible products priced above the target in ascending order and take the prices at approximately the one-third and two-thirds positions, respectively. If no eligible product is more expensive, we use the target price. We then round upward to a multiple of the increment assigned to the price range (\$1, \$5, \$10, \$50, or \$100), ensuring that the budget exceeds the target price.
\end{enumerate}
Finally, we check for conflicts among role backgrounds, task descriptions, product labels, and target products, and adjust them accordingly.

For example, a shopper may seek ``one sculptural seat'' with a dark or contrasting frame that suits a styled room, while avoiding bulky sofa modules.

During interaction, a disclosure policy guides how the user communicates the goal. The user begins by describing a broad product category, the intended use, or both, and gradually adds budget information, unwanted product features, and more specific needs, with at most one new key detail per turn.

When a recommendation is made, an acceptance policy guides the user's willingness to consider alternatives and the explanation needed before accepting. It also guides responses to unsuitable suggestions, such as giving a reason for rejection, asking for clarification, or comparing alternatives. Purchase decisions follow the shopping needs and preferences within the task's budget limit.

\FloatBarrier

\section{Runtime and Interaction}
\label{app:implementation-details}

\subsection{Runtime}
\label{app:unrealcv-runtime}
\label{app:user-embodiment}

\begin{table}[!t]
\centering

\caption{UnrealCV runtime interfaces.}
\setlength{\tabcolsep}{3pt}
\renewcommand{\arraystretch}{1}
{\small
\begin{tabularx}{\linewidth}{@{}l>{\raggedright\arraybackslash}X@{}}
\toprule
\textbf{Interface} & \textbf{Function} \\
\midrule
Connection
& Maintains communication between the runner and the UE5 scene. \\
User Pose
& Initializes and updates the embodied user's pose. \\
Action Execution
& Executes predefined UE5 actions for movement, view, and interaction. \\
State Query
& Queries object positions and orientations in the scene. \\
Camera Pose
& Aligns the first-person camera with the user's position and viewing direction. \\
Visual Capture
& Captures RGB, depth, and object-mask observations. \\
\bottomrule
\end{tabularx}
}
\label{tab:unrealcv-interface}
\end{table}

Table~\ref{tab:unrealcv-interface} summarizes the runtime interfaces used for environment control and observation capture.

As in SimWorld~\citep{ye2025simworld}, we use UnrealCV~\citep{DBLP:conf/mm/QiuZZQXKW17} for communication between the Python runtime and Unreal Engine. Each turn, the runtime validates the user's tool action and invokes the corresponding UE5 primitive. It then updates the first-person viewpoint to match the user's pose and captures the next observation.

Unreal Engine (UE) 5~\citep{ue5} provides collision handling, locomotion, and skeletal animation for the embodied user.
We combine UE Marketplace characters with custom Blueprints and animations to support callable actions such as turning, walking, and picking up items.
These actions update the physical scene and the observations used in subsequent dialogue.

\FloatBarrier

\subsection{Observations and Actions}
\label{app:observation-interface}
\label{app:action-space}

This section details the rendering, marker mappings, and access rules for the observations defined in Section~\ref{sec:observation_action}. Drawing on BASIS~\citep{zhang2025bootstrapping}, we use visual markers to identify interaction targets.

Before each run, we assign predefined RGB segmentation colors to the corresponding scene objects in UE5. Each distinct product and building has a fixed, unique color, while instances of the same product share that color. These colors identify products and buildings in the segmentation image and determine where their markers are overlaid on the RGB view.

During interaction, the runtime detects regions matching these colors in the current segmentation image and overlays contours or boxes with marker labels on the first-person RGB view. Depth determines the distance range within which these annotations are rendered. Adjacent instances of the same product share a contour, while separated instances receive separate contours. All regions of the same product share one label. Buildings receive boxes around their overall image regions.

The two roles use different marker-label mappings, and neither may disclose marker labels in dialogue. Users see numeric product labels, letter-based fixture/store labels, and area labels such as \texttt{A1}, while assistants see letter-based or extended alphanumeric product labels, numeric fixture/store labels, and area labels such as \texttt{1a}. Fixture, store, and area labels remain fixed according to each role's minimap mapping, while product labels are assigned sequentially within each role's current observation.

The assistant's \texttt{expected\_user\_action} uses its own role-specific handles, while the user's actual tool calls use the user's handles. The runtime maps both sets of handles to the corresponding underlying targets.

The current observation and physical state jointly determine which actions are executable, and actionable targets must appear in the corresponding visible-entity or interaction-state fields of the current observation.

Before inspection, the user receives a product's display name and temporary handle. In the Supermarket, picking up the product or asking the assistant confirms its price, flavor, packaging, allergens, and nutrition. In the Home Furnishings Store, \texttt{check\_item} or the assistant provides its price and catalog attributes, such as material, color, size category, and style.

Each assistant response contains a user-facing natural-language utterance together with a hidden one-element \texttt{assistant\_actions} array, an internal candidate ranking, and a required top-level \texttt{expected\_user\_action}. The assistant's \texttt{tool\_calls} array is always empty.
\paragraph{Assistant Dialogue Actions.}
\begingroup\raggedright The ten dialogue actions are \mbox{\texttt{Greet}}, \mbox{\texttt{AskPreference}}, \mbox{\texttt{ClarifyNeed}}, \mbox{\texttt{Navigate}}, \mbox{\texttt{Recommend}}, \mbox{\texttt{Explain}}, \mbox{\texttt{AnswerQuestion}}, \mbox{\texttt{Compare}}, \mbox{\texttt{HandleRejection}}, \mbox{\texttt{ConfirmPurchase}}.\par\endgroup

Each assistant turn contains exactly one primary action in the one-element \path{assistant_actions} array. All actions other than \texttt{Recommend} and \texttt{Navigate} require only a \texttt{type}. In both scenes, \texttt{Recommend} also requires \path{display_name}, \path{visible_now}, and \texttt{reason}. Supermarket \texttt{Navigate} requires \texttt{type}, \path{navigation_goal}, and \texttt{reason}; Home Furnishings Store additionally requires \path{next_waypoint} and \path{target_approach}.

The final navigation goal is a fixture in Supermarket or a store or display area in Home Furnishings Store. It may be unseen, but the immediate navigation target must be visible.

The runtime prompt templates are collected together in Appendix~\ref{app:prompt-templates}. The released benchmark includes the complete assembled prompts.

\newcommand{\assistantprompttemplate}{%
\begin{tcolorbox}[
        title=\textbf{Prompt Template for Recommendation Assistant},
        colframe=blue!50!black,
        colback=blue!5!white,
        coltitle=white,
        fonttitle=\small\bfseries,
        arc=1mm,
        boxsep=2pt,
        fontupper=\small,
        before upper={\raggedright},
        before skip=\baselineskip,
        after skip=\baselineskip,
        breakable,
        lines before break=4,
        title after break={\textbf{Prompt Template for Recommendation Assistant (Continued)}},
        enhanced
]
        You are a sales assistant in the active UE5 shopping scene (Supermarket or Home Furnishings Store).
        Understand the user's needs through natural conversation and help them find, compare, and buy
        suitable products.

        \textbf{Available Context.}
        Use the active scene's environment and action reference, catalog, location index, and interaction history.
        Images show only the current observation: the layout map and your own RGB overlay.

        \textbf{Generation Order.}
        Use expressed needs, current observations, dialogue history, and catalog facts to generate
        \promptcode{recommendation_ranking} with ten distinct products. Then select one primary
        \promptcode{assistant_actions} entry and one \promptcode{expected_user_action}, and finally generate
        natural \promptcode{content}. Rank 1 need not be recommended in the current turn.

        \textbf{Interaction Strategy.}
        Begin with a greeting and a question about the need. Address the user's latest question, concern,
        or rejection reason and advance one step per turn in one or two natural sentences. Ask when missing
        information affects the next choice; otherwise continue navigation, inspection, or recommendation.
        Recommend when asked or when a visible item is very likely to fit. Check uncertain details and
        compare when requested. Use expressed preferences and update suggestions after rejection.

        \textbf{Scene Grounding.}
        Give directions from the user's current position and facing, indicated by the red map arrow.
        The final goal may be unseen, but the target used by the immediate action must be actionable in
        the current observation. Use a visible waypoint or a movement or view adjustment to reveal an
        unseen target. In Home Furnishings Store, locate an unseen storefront by turning before navigating to it.
        Explore locally after arrival. Use the new observation after movement and prioritize it over history.
        Only products in \promptcode{visible_items} are currently visible; the location index supports route
        planning. Ground factual claims in the catalog, observations, and dialogue history.

        \textbf{Output Boundary.}
        The user hears only \promptcode{content}; rankings, action fields, and private handles remain
        internal. Keep tool names and raw catalog labels out of speech. Your \promptcode{tool_calls} is empty;
        the runtime executes the user's chosen action. During item confirmation, the expected action is
        \promptcode{add_to_cart} or the scene's decline action: \promptcode{drop_item} in Supermarket or
        \promptcode{do_not_add_to_cart} in Home Furnishings Store.

\begin{minipage}{\linewidth}
\textbf{Allowed Assistant Action Labels.}
        \begin{tcolorbox}[colback=white, colframe=gray!30, arc=0mm, boxsep=0pt, left=4pt, right=4pt, top=2pt, bottom=2pt]
            \ttfamily\small\raggedright
            Greet, AskPreference, ClarifyNeed, Navigate, Recommend, Explain,\par
            AnswerQuestion, Compare, HandleRejection, ConfirmPurchase
        \end{tcolorbox}
\end{minipage}

        Record exactly one label. All actions require \promptcode{type}. \promptcode{Navigate} requires
        \promptcode{navigation_goal} and \promptcode{reason}; Home Furnishings Store also requires
        \promptcode{next_waypoint} and \promptcode{target_approach}. \promptcode{Recommend} requires
        \promptcode{display_name}, \promptcode{visible_now}, and \promptcode{reason}. The final navigation goal
        may be unseen, but the immediate expected action must be executable. Use a fixture handle in Supermarket
        and a store or area handle in Home Furnishings Store, or \promptcode{null} if the goal is unknown.
        In Home Furnishings Store, set \promptcode{next_waypoint} to a visible store or area handle only for
        \promptcode{move_to_store} or \promptcode{move_to_area}; otherwise use \promptcode{null}.
        Set \promptcode{target_approach} to \promptcode{inside_doorway} or \promptcode{outside_doorway}
        only for \promptcode{move_to_store}; otherwise use \promptcode{null}.

        \begin{minipage}{\linewidth}
\textbf{Inputs Available to the Assistant:}
        \begin{tcolorbox}[colback=white, colframe=gray!30, arc=0mm, boxsep=0pt, left=4pt, right=4pt, top=2pt, bottom=2pt]
            \ttfamily\small\raggedright
            \promptslot{current observation: visible items, scene-specific targets, navigation and shopping state}\par
            \promptslot{store layout map image}\par
            \promptslot{current RGB overlay with private assistant handles}\par
            \promptslot{full catalog: product names, prices, and public labels}\par
            \promptslot{product-location index: fixture handles or store names with area handles}\par
            \promptslot{conversation history and action history}
        \end{tcolorbox}
\end{minipage}

        \begin{minipage}{\linewidth}
\textbf{Required Output Format:}
        \begin{tcolorbox}[colback=white, colframe=gray!30, arc=0mm, boxsep=0pt, left=4pt, right=4pt, top=2pt, bottom=2pt]
            \ttfamily\small\raggedright
            \{"type":"final",\par
            \ \ "recommendation\_ranking":[\{"rank":1,\par
            \ \ "display\_name":"...","reason":"...",\par
            \ \ "evidence\_from\_dialogue":["..."],"confidence":0.0\}],\par
            \ \ "assistant\_actions":[\{"type":"AskPreference"\}],\par
            \ \ "expected\_user\_action":\{"name":"idle","arguments":\{\}\},\par
            \ \ "content":"...","tool\_calls":[]\}
        \end{tcolorbox}
\end{minipage}

        Output one JSON object with non-empty \promptcode{content}, no Markdown, and empty \promptcode{tool_calls}.
        The example abbreviates the ten-entry ranking; use ranks 1--10. Assess confidence independently in
        $[0,1]$, with lower confidence for weaker plausible matches. The expected action uses the active scene's
        user tool schema and assistant-side actionable handles; use \promptcode{idle} when no physical action
        is appropriate outside confirmation.

\end{tcolorbox}%
}

\noindent\begin{minipage}{\linewidth}
\paragraph{User Embodied Tool Actions.}
Each scene uses the shared tools plus its scene-specific tools.
\par\smallskip\begingroup
\setlength{\tabcolsep}{5pt}
\renewcommand{\arraystretch}{1.2}
\noindent{\small
\begin{tabularx}{\linewidth}{@{}l>{\raggedright\arraybackslash}X@{}}
\textbf{Shared} & \mbox{\texttt{move\_forward}}, \mbox{\texttt{move\_backward}}, \mbox{\texttt{turn\_and\_move}}, \mbox{\texttt{move\_and\_turn}}, \mbox{\texttt{side\_step}}, \mbox{\texttt{turn}}, \mbox{\texttt{look\_up\_down}}, \mbox{\texttt{add\_to\_cart}}, \mbox{\texttt{idle}} \\
\addlinespace[5pt]
\textbf{Supermarket} & \mbox{\texttt{move\_to\_fixture}}, \mbox{\texttt{pick\_up\_item}}, \mbox{\texttt{drop\_item}} \\
\addlinespace[5pt]
\textbf{Home Furnishings Store} & \mbox{\texttt{move\_to\_store}}, \mbox{\texttt{move\_to\_area}}, \mbox{\texttt{check\_item}}, \mbox{\texttt{do\_not\_add\_to\_cart}} \\
\end{tabularx}
}\endgroup
\end{minipage}\par\smallskip
The Supermarket target action navigates to a currently visible fixture or shelf. Its \texttt{nearest} approach
moves to the closest reachable point without an extra facing adjustment; \texttt{front} and \texttt{back} move to
fixed side-specific viewpoints and align facing, and are allowed only when the current
\texttt{visible\_fixture\_nav} mapping supports that side. Home Furnishings Store instead supports navigation to a visible storefront doorway or display area. Its \texttt{move\_to\_store} action includes an \texttt{approach} argument with values \texttt{outside\_doorway} and \texttt{inside\_doorway}; \texttt{move\_to\_area} moves to a fixed point within a currently visible display area.

\paragraph{User Tool-Action Constraints.}
Each user turn contains exactly one tool action, constrained by the current observation and physical state. Object interactions and navigation targets use identifiers from the user's current observation. User and assistant identifiers are private to each role and mapped by the runtime to the same underlying targets. Additional state-dependent constraints are specified in the prompt templates below. Executed actions are recorded in the dialogue history to inform subsequent assistant responses.

\newcommand{\userprompttemplate}{%
\begin{tcolorbox}[
        title=\textbf{Prompt Template for User Simulator},
        colframe=green!45!black,
        colback=green!5!white,
        coltitle=white,
        fonttitle=\small\bfseries,
        arc=1mm,
        boxsep=2pt,
        fontupper=\small,
        before upper={\raggedright},
        before skip=\baselineskip,
        after skip=\baselineskip,
        breakable,
        lines before break=4,
        title after break={\textbf{Prompt Template for User Simulator (Continued)}},
        enhanced
]
        Simulate a real user shopping from a first-person view in the active UE5 scene
        (Supermarket or Home Furnishings Store). Speak naturally and stay consistent with your private shopping goal.

        \textbf{Available Context.}
        Use your private profile and shopping goal, the active scene's action reference, and the interaction history.
        You receive the current text observation and your own RGB overlay, without a layout map or past images.

        \textbf{Behavior and Speech.}
        Use the profile, private goal, current observation, and interaction history to select exactly one
        tool action, then generate a consistent natural reply. Explicit behavior parameters guide interaction
        and decisions, with Big Five traits as secondary tendencies. Use only the supplied profile attributes.

        \textbf{Disclosure.}
        Follow \promptcode{disclosure_policy} to reveal the product category, use context, budget, and
        preferences progressively, adding at most one new key detail per turn.

        \textbf{Choice and Purchase.}
        Keep the private goal, preferences, and budget fixed. Follow \promptcode{acceptance_policy} for
        substitute tolerance, explanation needs, and rejection style. Assess suggestions against the current
        view and your needs, then follow, ask, reject, or explore. Before inspection, visible products provide
        names and handles only. Confirm details by asking, using \promptcode{pick_up_item} in Supermarket,
        or \promptcode{check_item} in Home Furnishings Store. Buy when the item meets the key need, budget, and avoid constraints.

        \textbf{Output and State.}
        Put natural speech in \promptcode{content} and exactly one action in \promptcode{tool_calls}, using
        current user-side handles. Keep handles and tool names out of speech. Use \promptcode{idle} for
        conversation without a physical action outside confirmation. Once holding or checking an item, choose
        \promptcode{add_to_cart} or the scene's decline action: \promptcode{drop_item} in Supermarket or
        \promptcode{do_not_add_to_cart} in Home Furnishings Store. If unsure, decline and explain or ask in the same turn.

\begin{minipage}{\linewidth}
\textbf{Private and Runtime Inputs:}
        \begin{tcolorbox}[colback=white, colframe=gray!30, arc=0mm, boxsep=0pt, left=4pt, right=4pt, top=2pt, bottom=2pt]
            \ttfamily\small\raggedright
            \promptslot{private shopping goal, preferences, and budget}\par
            \promptslot{filtered user background and life context}\par
            \promptslot{Big Five traits and interaction parameters}\par
            \promptslot{disclosure and acceptance policies}\par
            \promptslot{current observation: visible items, scene-specific targets, and navigation}\par
            \promptslot{current RGB overlay with private user handles}\par
            \promptslot{held-item or inspection state, and cart contents}\par
            \promptslot{conversation history and latest assistant response}
        \end{tcolorbox}
\end{minipage}

        \begin{minipage}{\linewidth}
\textbf{Allowed Tools by Scene:}
        \begin{tcolorbox}[colback=white, colframe=gray!30, arc=0mm, boxsep=0pt, left=4pt, right=4pt, top=2pt, bottom=2pt]
            \ttfamily\small\raggedright
            \textnormal{Shared:}\par
            move\_forward: \{"duration": number\}\par
            move\_backward: \{"duration": number\}\par
            turn\_and\_move: \{"direction": "right|left", "degree": number, "duration": number\}\par
            move\_and\_turn: \{"direction": "right|left", "degree": number, "duration": number\}\par
            side\_step: \{"direction": "left|right", "duration": number\}\par
            turn: \{"direction": "right|left", "degree": number\}\par
            look\_up\_down: \{"direction": "up|down", "degree": number\}\par
            add\_to\_cart: \{"item\_id": "string"\}\par
            idle: \{\}\par
            \textnormal{Supermarket Only:}\par
            move\_to\_fixture: \{"fixture\_id": "string", "approach": "nearest|front|back"\}\par
            pick\_up\_item: \{"item\_id": "string"\}\par
            drop\_item: \{\}\par
            \textnormal{Home Furnishings Store Only:}\par
            move\_to\_store: \{"store\_id": "string",\par
            \ \ "approach": "outside\_doorway|inside\_doorway"\}\par
            move\_to\_area: \{"area\_id": "string"\}\par
            check\_item: \{"item\_id": "string"\}\par
            do\_not\_add\_to\_cart: \{\}
        \end{tcolorbox}
\end{minipage}

        \begin{minipage}{\linewidth}
\textbf{Required Output Format:}
        \begin{tcolorbox}[colback=white, colframe=gray!30, arc=0mm, boxsep=0pt, left=4pt, right=4pt, top=2pt, bottom=2pt]
            \ttfamily\small\raggedright
            \{"type":"final","content":"...",\par
            \ \ "tool\_calls":[\{"name":"idle","arguments":\{\}\}]\}
        \end{tcolorbox}
\end{minipage}

        Return one valid JSON object with non-empty \promptcode{content} and exactly one action in
        \promptcode{tool_calls}. Use only shared tools and the active scene's additions, with object arguments
        and this turn's user-side handles. Do not output Markdown or additional text.

\end{tcolorbox}%
}

\paragraph{Scene-Specific Prompt Adaptation.}
\label{app:scene-specific-prompts}

The scenes share role policies, handle rules, grounding requirements, and termination logic. Their prompts differ in catalogs, spatial entities, and interaction schemas. The product-location index links Supermarket products to fixtures and Home Furnishings Store products to stores and display areas, supporting the assistant's route planning.

In Home Furnishings Store, targeted navigation uses only the current role's visible store or area handles.
\texttt{inside\_doorway} and \texttt{outside\_doorway} specify fixed points just inside and outside the same store entrance, respectively. For longer-distance navigation, the outside point of a visible store can serve as an intermediate waypoint. If the storefront is not visible, turn to find it rather than taking medium or long
free movements. After reaching an area, explore locally instead of repeatedly navigating to the same area.
The assistant's final navigation goal may be unseen; its immediate waypoint must be visible.

\FloatBarrier

\subsection{Interaction and Termination}
\label{app:interaction-protocol}
\label{app:moderator-termination}

We coordinate assistant--user turns with ChatArena~\citep{ChatArena}, which TopDial~\citep{wang2023target} uses for role-playing dialogue simulation. Profile--goal pairs follow task-bank order. After each exchange, the runtime executes and records the user's action, checks termination, and queries the moderator if the episode remains active. The conversation judge scores the completed trajectory.

Each assistant turn records an internal Top-10 ranking and one dialogue-action annotation. A spoken recommendation of a specific item requires a \texttt{Recommend} annotation and counts as one recommendation attempt. These fields remain hidden from the user.

Episodes may end through purchase completion, interaction limits, dialogue closure, or motion-stall conditions:
\begin{enumerate}
\item \textbf{Purchase Completion.} The user adds an item to the cart. This outcome takes precedence over the turn and recommendation limits.
\item \textbf{Turn Limit.} The episode reaches the standard limit of 20 turns.
\item \textbf{Recommendation Exhaustion.} After the user's response and action, at least three explicit recommendations have been recorded without a purchase. Attempts are counted from \texttt{Recommend} annotations and may span nonconsecutive turns, with or without explicit rejection. The per-turn Top-10 ranking is recorded separately and does not contribute to this count.
\item \textbf{Moderator Termination.} If the episode remains active after runtime checks, the moderator uses the conversation history, latest assistant and user messages, and current user action to decide whether to end the interaction. It returns \texttt{yes} when the user clearly gives up, declines further help, or ends the interaction, or when both sides have completed purchase confirmation or closing conversation. It returns \texttt{no} while the user continues clarifying needs, exploring, comparing or inspecting products, or seeking help after rejection or navigation difficulty.
\item \textbf{Motion Stall.} Three consecutive translational actions with negligible displacement are treated as persistent obstruction in the physical environment and end the episode.
\end{enumerate}
The moderator prompt template is provided in Appendix~\ref{app:prompt-templates}.

\newcommand{\moderatorprompttemplate}{%
\begin{tcolorbox}[
        title=\textbf{Prompt Template for Dialogue Moderator},
        colframe=purple!55!black,
        colback=purple!5!white,
        coltitle=white,
        fonttitle=\small\bfseries,
        arc=1mm,
        boxsep=2pt,
        fontupper=\small,
        before upper={\raggedright},
        before skip=\baselineskip,
        after skip=\baselineskip,
        breakable,
        lines before break=4,
        title after break={\textbf{Prompt Template for Dialogue Moderator (Continued)}},
        enhanced
]
        Decide whether the shopping conversation between the assistant and the user should end.

        \textbf{Available Context.}
        Use the dialogue and action history and the current turn. You receive no images, observation snapshot,
        private user profile, or shopping goal.

        \textbf{Decision Policy.}
        Use the history and latest user action to decide whether to end, continuing by default.
        Output \promptcode{yes} for a completed cart addition, an explicit wish to stop, or a natural close
        to the shopping interaction. Output \promptcode{no} while needs are being clarified or products are
        being sought, inspected, compared, or considered, including after a rejection when the user
        remains willing to continue. Output only one lowercase word, \promptcode{yes} or \promptcode{no}.

\begin{minipage}{\linewidth}
\textbf{Inputs Available to the Moderator:}
        \begin{tcolorbox}[colback=white, colframe=gray!30, arc=0mm, boxsep=0pt, left=4pt, right=4pt, top=2pt, bottom=2pt]
            \ttfamily\small\raggedright
            \promptslot{compact conversation history}\par
            \promptslot{current turn: latest assistant and user dialogue and user action}\par
            \promptslot{explicit yes/no termination question}
        \end{tcolorbox}
\end{minipage}

        \textbf{Required Output Format:} \promptcode{yes} or \promptcode{no}.
\end{tcolorbox}%
}

\newcommand{\judgeprompttemplate}{%
\begin{tcolorbox}[
        title=\textbf{Prompt Template for Conversation Judge},
        colframe=orange!60!black,
        colback=orange!5!white,
        coltitle=white,
        fonttitle=\small\bfseries,
        arc=1mm,
        boxsep=2pt,
        fontupper=\small,
        before upper={\raggedright},
        before skip=\baselineskip,
        after skip=\baselineskip,
        breakable,
        lines before break=4,
        title after break={\textbf{Prompt Template for Conversation Judge (Continued)}},
        enhanced
]
        You are an evaluator. Assess the complete interaction in the active scene (Supermarket or Home Furnishings Store)
        from the user's point of view, using the dimensions listed below.
        Use user context, text observations, dialogue, actions, and the final state; no images are supplied.
        Product details become available through pickup in Supermarket or inspection in Home Furnishings Store;
        the user then chooses whether to add the item to the cart. A cart addition alone does not establish
        that the purchase meets the user's needs.

        \textbf{Evaluation Policy.}
        Read text observations, dialogue, and user actions in time order, using later observations
        and the final state to evaluate the complete interaction. Judge Preference Alignment against expressed
        needs and Overall Satisfaction against the complete shopping need and interaction experience.

        Use basic product knowledge and check claims against verifiable scene, product, and state evidence,
        taking account of the assistant's catalog access. Assess directions and distances only to the extent
        supported by text observations. When an action fails, consider the assistant's guidance, the user's
        actual action, and the assistant's subsequent adjustment.

\textbf{Survey Dimensions and Scale.}
        Assign an integer from 1 (very poor) to 5 (excellent) for each dimension:
        \promptcode{informativeness}, \promptcode{naturalness},
        \promptcode{groundedness}, \promptcode{preference_alignment}, \promptcode{proactiveness},
        \promptcode{non_repetition}, and \promptcode{overall_satisfaction}. Provide a brief English rationale
        for each, an English summary, and any issues. The assessment questions are given in
        Appendix~\ref{app:communication-quality}.

        \begin{minipage}{\linewidth}
\textbf{Inputs Available to the Judge:}
        \begin{tcolorbox}[colback=white, colframe=gray!30, arc=0mm, boxsep=0pt, left=4pt, right=4pt, top=2pt, bottom=2pt]
            \ttfamily\small\raggedright
            \promptslot{user context: filtered profile, shopping need, preferences, and budget}\par
            \promptslot{interaction history: text observations, dialogue, and actions in turn order}\par
            \promptslot{final user state: ending, cart, held-item or inspection state}
        \end{tcolorbox}
\end{minipage}

        \begin{minipage}{\linewidth}
\textbf{Required Output Format:}
        \begin{tcolorbox}[colback=white, colframe=gray!30, arc=0mm, boxsep=0pt, left=4pt, right=4pt, top=2pt, bottom=2pt]
            \ttfamily\small\raggedright
            \{"scores":\{"informativeness":1, ... ,"overall\_satisfaction":1\},\par
            \ \ "rationales":\{"informativeness":"...", ...\},\par
            \ \ "summary":"...","issues":[]\}
        \end{tcolorbox}
\end{minipage}

        Return one valid JSON object with all listed keys in both \promptcode{scores} and
        \promptcode{rationales}; the schema above is abbreviated. Use an empty \promptcode{issues} array
        when there are no issues. Do not output Markdown or extra text.
\end{tcolorbox}%
}

\FloatBarrier

\section{Evaluation Methods}
\label{app:evaluation-details}
\subsection{Metric Definitions}
\label{app:metric-definitions}
The main evaluation metrics follow four dimensions: Overall Outcome, Dialogue Quality, Recommendation Utility, and Navigation Effectiveness. We first specify counting and aggregation rules shared by these metrics.

A turn is counted if a record for that turn exists. Let $\mathcal{R}_e$ be the set of distinct recorded interaction turns in episode $e$; its total turn count is $T_e=|\mathcal{R}_e|$. For GPT-5.6 Sol-user results with SDs, we divide each environment's 100 tasks into four disjoint subsets of 25, each covering all 25 profiles. Using seed 42, we shuffle the four tasks for each profile and assign one to each subset. All assistants and ablation conditions use this partition.

For each assistant and metric, we first calculate the mean $\mu_g$ over eligible episodes in each subset $g$, using the table's units. We then report the average of the four subset means, $\bar{\mu}=\tfrac14\sum_{g=1}^{4}\mu_g$, with the sample SD of those means, $s=\sqrt{\sum_{g=1}^{4}(\mu_g-\bar{\mu})^2/3}$, shown as a subscript. GPT-4o-user results report cohort means. Proportions reported without SDs use the full cohort or their stated conditional denominator. Event-pooled statistics use the relevant event counts as their denominators.

\subsubsection*{Overall Outcome}
The cumulative cart-addition rate at turn $t$ is
\[\mathrm{SR}(t)=\frac{100}{N}\sum_e\mathbf{1}[C_e=1\land T_e\le t].\]
\paragraph{Success Rate (SR) and Turn.}
For both environments, $C_e=1$ when the final cart is nonempty. Within a cohort or main-table subset of $N$ episodes, $\mathrm{SR}=100N^{-1}\sum_e C_e$ and $\overline{T}=N^{-1}\sum_e T_e$. Both statistics include all $N$ episodes. Cart addition ends successful episodes, so $T_e$ is also the cart-addition turn for those episodes. Each main-table assistant/environment cohort contains 100 episodes, with $N=25$ per subset; each available GPT-4o-user cohort contains $N=25$; each ablation condition contains $N=100$. The sample sizes above give the denominators for SR, T, R@5 and NDCG@10. Navigation metrics are aggregated as described below.

\subsubsection*{Dialogue Quality}
\paragraph{Subjective Ratings.}\label{app:communication-quality}
Dialogue quality is reported on the following seven anchored dimensions on a 1--5 scale.

\begingroup
\textbf{Informativeness.} When the user needs information or guidance, does the Assistant provide specific and relevant content that supports a decision or next action rather than vague statements?

\textbf{Naturalness.} Are the Assistant's wording, tone, and conversational pacing natural, concise, and coherent, consistent with everyday shopping conversation and basic common sense, like a real shopping assistant, without system jargon or formulaic phrasing?

\textbf{Groundedness.} Are the Assistant's claims consistent with the available scene, product, state, and dialogue evidence, without unsupported assertions, contradictions, or treating information unavailable to the user as something the user already knows?

\textbf{Preference Alignment.} Does the Assistant consistently respect the user's stated needs, budget, preferences, and constraints, and adapt after rejection or hesitation?

\textbf{Proactiveness.} When reasonable opportunities to progress occur, do the Assistant's questions, navigation guidance, comparisons, recommendations, or confirmations move the user closer to completing the shopping task rather than causing stalls or aimless actions?

\textbf{Non-repetition.} Does the Assistant avoid repeating the same question, route, product, or wording without new justification, and change strategy when progress continues to stall?

\textbf{Overall Satisfaction.} Considering need fulfillment, process efficiency, trustworthiness, and the overall experience, how satisfied are you with the Assistant's performance across the complete interaction?
\par\endgroup

\paragraph{Human-Evaluation Agreement.}
Pearson's $r$ measures linear association between paired assistant-level means, and Spearman's $\rho$ measures rank agreement. These coefficients support human validation of task metrics and dialogue ratings; definitions and comparison procedures are given in Appendix~\ref{app:human-validation}.

\subsubsection*{Recommendation Utility}
\paragraph{Final-Cart Ranking (R@5 and NDCG@10).}
R@5 and NDCG@10 evaluate each saved recommendation list against the products the user eventually purchases. R@5 measures the fraction of final-cart products appearing in the top five recommendations. NDCG@10 also considers their positions in the top ten. Let $P_e$ be the final-cart product set, $M_e$ the number of saved recommendation lists, and $R_{et}$ the ranked list at turn $t$. Ranks follow list order; repeated products receive credit at their first occurrence without shifting subsequent positions. Define
\[q_{et}=\frac{|\operatorname{unique}(R_{et}[1:5])\cap P_e|}{|P_e|},\qquad
z_{et}=\frac{\sum_{r=1}^{10}\mathrm{rel}_{etr}/\log_2(r+1)}{\sum_{r=1}^{\min(10,|P_e|)}1/\log_2(r+1)}.\]
These expressions apply when $P_e$ is nonempty; $\mathrm{rel}_{etr}$ is one only for the first occurrence of a final-cart product. For either score $x\in\{q,z\}$, set
\[s_e(x)=\begin{cases}M_e^{-1}\sum_{t=1}^{M_e}x_{et},&|P_e|>0\ \text{and}\ M_e>0,\\0,&\text{otherwise}.\end{cases}\]
R@5 and NDCG@10 measure agreement between recommendation rankings and final-cart items and are reported as percentages:
\[\mathrm{R@5}=\frac{100}{N}\sum_e s_e(q),\qquad
\mathrm{NDCG@10}=\frac{100}{N}\sum_e s_e(z).\] We first average the recommendation-list scores within each episode, then average the episode scores over all episodes in the corresponding subset or cohort. An empty recommendation list scores zero. An episode with an empty final cart or no recommendation records also scores zero. 

\subsubsection*{Navigation Effectiveness}
\paragraph{Action Following Rate (AFR).}
AFR measures how often the user selects the action type suggested by the assistant. For each saved navigation event, we compare the assistant's structured expected user action with the user's first tool call in that turn. Let $V_e$ contain events whose expected tool name is in the environment's legal set. Then
\[\mathrm{AFR}_e=\frac{\sum_{t\in V_e}\mathbf{1}[\widehat a_{et}=a_{et}]}{|V_e|}.\]
\begingroup\raggedright
The complete legal action sets are:
\textbf{Supermarket (12):} \texttt{move\_forward}, \texttt{move\_backward}, \texttt{turn\_and\_move}, \texttt{move\_and\_turn}, \texttt{side\_step}, \texttt{turn}, \texttt{look\_up\_down}, \texttt{move\_to\_fixture}, \texttt{pick\_up\_item}, \texttt{add\_to\_cart}, \texttt{drop\_item}, and \texttt{idle}. \textbf{Home Furnishings Store (13):} \texttt{move\_forward}, \texttt{move\_backward}, \texttt{turn\_and\_move}, \texttt{move\_and\_turn}, \texttt{side\_step}, \texttt{turn}, \texttt{look\_up\_down}, \texttt{move\_to\_store}, \texttt{move\_to\_area}, \texttt{check\_item}, \texttt{add\_to\_cart}, \texttt{do\_not\_add\_to\_cart}, and \texttt{idle}.\par
\endgroup

AFR measures tool-name agreement over the legal actions above, independently of arguments and execution outcomes. An absent or different actual action scores zero; missing or illegal expected actions are excluded from $V_e$.
\paragraph{Local Navigation Alignment (LNSR).}
LNSR measures agreement in both action type and local navigation parameters. Let $D_e\subseteq V_e$ contain events whose expected local parameters can be compared under the rules below, and let $b_{et}$ indicate whether those parameters match. We compute
\[\mathrm{LNSR}_{e}=\frac{\sum_{t\in D_e}\mathbf{1}[\widehat a_{et}=a_{et}]b_{et}}{|D_e|}.\]
\begingroup\raggedright
Supermarket includes six actions: \texttt{turn\_and\_move}, \texttt{move\_and\_turn}, \texttt{side\_step}, \texttt{turn}, \texttt{look\_up\_down}, and \texttt{move\_to\_fixture}. Home Furnishings Store includes seven: \texttt{turn\_and\_move}, \texttt{move\_and\_turn}, \texttt{side\_step}, \texttt{turn}, \texttt{look\_up\_down}, \texttt{move\_to\_store}, and \texttt{move\_to\_area}.\par
\endgroup
The five directional actions compare whether the expected and actual directions match (left, right, up or down). Angle magnitude, distance and duration are not evaluated. For \texttt{move\_to\_fixture}, only target identity is compared. For \texttt{move\_to\_store} and \texttt{move\_to\_area}, an expected location is required; \texttt{move\_to\_store} additionally compares \texttt{approach} when the expected action supplies it. Target matching resolves the two roles' different marker labels and checks whether the expected and actual actions refer to the same object or location. Each event requires both tool and parameter matches; a comparable expected action with a missing or mismatched actual tool or parameter scores zero. Events with unavailable expected targets or directions are excluded from $D_e$.
\paragraph{Navigation Aggregation.}
We first calculate AFR and LNSR for each episode. For each metric, we average the scores over episodes containing at least one eligible event and express the result as a percentage. Let $E_m$ denote these episodes for metric $m$. The mean within each subset or cohort is
\[100|E_m|^{-1}\sum_{e\in E_m}m_e.\]
Subset results are then combined using the aggregation rules above.
\paragraph{Cumulative Goal Visibility (CGV).}
CGV tracks the visibility of the destination recorded in the assistant's \texttt{navigation\_goal}: a fixture in Supermarket or a store or display area in Home Furnishings Store.

For each recorded navigation goal with a checkable current observation, we examine the current observation and the following $h$ turns. The goal counts as visible once it appears in any saved observation within this window. In Home Furnishings Store, a visible product associated with the destination store or area also establishes goal visibility. We pool eligible navigation-goal events and use the same denominator for every horizon.

Figures report the increase over current visibility, $\Delta\mathrm{CGV}_h=\mathrm{CGV}_h-\mathrm{CGV}_0$, in percentage points for $h=1,\ldots,5$ and End. End extends the window through the saved episode end.

\subsubsection*{Supplementary Metrics}

For each task, we construct a fixed reference item set $A_e$ containing products that meet the task's shopping requirements and budget.

\paragraph{Reference Purchase Rate (RPR).}
RPR is the proportion of all episodes in which the purchased item belongs to $A_e$. RPR$\mid$Cart measures this proportion among episodes with a purchase.

\paragraph{Reference-item Mean Reciprocal Rank (R-MRR@10).}
Each saved recommendation list receives the reciprocal rank of its first reference item within the top ten, or zero if none appears. We average these scores first within each episode and then across episodes with at least one saved recommendation record in the corresponding subset or cohort.

\paragraph{Catalog Validity (Catalog \%).}
Catalog validity is the percentage of ranked entries that belong to the product catalog.

\subsection{Human Evaluation Protocol}
\label{app:conversation-judge}\label{app:human-validation}
Participants fill in a user profile. We randomly sample five tasks from the 100-task Supermarket bank (seed 42) and assign adjacent task pairs cyclically, with each task assigned to two participants. Each participant completes the two assigned tasks in sequence and interacts with all five assistants on each task. Models are presented under letter labels, and evaluation order varies across participants. Before each task, participants read the assigned shopping goal, then converse with the assistant and choose actions using the scene images.

After each interaction, participants complete the questionnaire using the guidance displayed in the interface. The LLM judge independently scores the same trajectories on the same dimensions to compare the two assessments of assistant performance. Records use coded participant identifiers and include profiles, dialogue, actions and questionnaire responses.

The human interactions provide a reference for assessing whether simulated users preserve relative assistant performance (Section~\ref{sec:human_evaluation}). We compare assistant-level means from the sampled human tasks and the full automatic-user cohort using the same main-table metric definitions and the same LLM judge for dialogue ratings (Table~\ref{tab:human-validation-supermarket}). Human means average all trajectories for each assistant, and automatic means are aggregated over the four profile-stratified subsets defined in Appendix~\ref{app:metric-definitions}. We quantify agreement using Pearson's $r$ and Spearman's $\rho$. Pearson's $r$ is centered covariance normalized by the product of standard deviations; Spearman's $\rho$ is Pearson correlation of average ranks, with ties assigned midranks. Both coefficients use unrounded means, and differences below numerical precision are treated as ties when computing ranks.

On the same human trajectories, we also examine how closely LLM judge ratings reflect participants' assessments. Figure~\ref{fig:human-questionnaire-judge-ratings} compares questionnaire and LLM judge ratings averaged over the 10 trajectories for each assistant. Sample sizes for both comparisons are reported in Section~\ref{sec:human_evaluation}.

\clearpage

\section{Supplementary Results}
\label{app:additional-results}\label{app:full-experimental-results}
We report GPT-4o-user results, purchase quality, failure patterns, goal visibility, and the ablation protocol. Metric definitions and eligible sample sizes are in Appendix~\ref{app:evaluation-details}.
\subsection{GPT-4o Users}\label{app:user-simulator-sensitivity}
To examine assistant performance with simulated users of different comprehension abilities, we additionally use GPT-4o as the user simulator to evaluate five assistant models in both Supermarket and Home Furnishings Store. In each environment, we select one task for each of the 25 user profiles. Each assistant completes all 25 tasks per environment, yielding 250 interaction trajectories in total.

With GPT-4o users, Gemini 3.1 Pro again leads purchase and recommendation metrics in both environments, while GPT-5.6 Sol retains the highest AFR. These results extend their respective strengths in purchase and recommendation performance and action alignment observed in the main experiments (Table~\ref{tab:user4o_results}).
\begin{table}[!t]
\centering

\setlength{\tabcolsep}{3pt}
\caption{GPT-4o-user results (25 tasks per assistant/environment), reported as means without SDs. Cart-ranking scores include non-purchases as zero; navigation uses eligible episodes.}
\label{tab:user4o_results}
{\small
\begin{tabular*}{\linewidth}{@{\extracolsep{\fill}}llcccccc@{}}
\toprule
\multirow{2}{*}{\textbf{Environment}} & \multirow{2}{*}{\textbf{Model}} & \multicolumn{2}{c}{\textbf{Outcome}} & \multicolumn{2}{c}{\textbf{Navigation}} & \multicolumn{2}{c}{\textbf{Recommendation}} \\
\cmidrule(lr){3-4}\cmidrule(lr){5-6}\cmidrule(lr){7-8}
 &  & \textbf{SR (\%)} & \textbf{Turn} & \textbf{AFR (\%)} & \textbf{LNSR (\%)} & \textbf{R@5 (\%)} & \textbf{NDCG@10 (\%)} \\
\midrule
\multirow{5}{*}{Supermarket} & Qwen3.6-Plus & 52.0 & 11.9 & 68.8 & 32.6 & 41.7 & 35.1 \\
 & GPT-4o & 68.0 & 9.9 & 68.6 & 15.0 & 35.1 & 30.5 \\
 & GPT-5.6 Sol & 72.0 & 10.7 & \textbf{89.1} & 32.3 & 59.0 & 50.0 \\
 & Claude Opus 5 & 76.0 & 10.8 & 72.7 & 42.7 & 51.6 & 46.3 \\
 & Gemini 3.1 Pro & \textbf{92.0} & 9.4 & 74.8 & \textbf{50.1} & \textbf{73.4} & \textbf{57.7} \\
\midrule
\multirow{5}{*}{\makecell[l]{Home\\Furnishings\\Store}} & GPT-4o & 16.0 & 6.3 & 61.9 & \textbf{75.1} & 8.9 & 8.0 \\
 & Claude Opus 5 & 32.0 & 10.9 & 64.5 & 57.0 & 21.7 & 19.9 \\
 & Qwen3.6-Plus & 60.0 & 10.0 & 63.2 & 51.5 & 39.4 & 36.5 \\
 & GPT-5.6 Sol & 72.0 & 10.4 & \textbf{71.6} & 72.4 & 52.6 & 47.8 \\
 & Gemini 3.1 Pro & \textbf{76.0} & 11.2 & 69.5 & 70.6 & \textbf{60.5} & \textbf{51.1} \\
\bottomrule
\end{tabular*}
}
\end{table}
\begin{figure}[!t]
\centering
\includegraphics[width=0.96\linewidth]{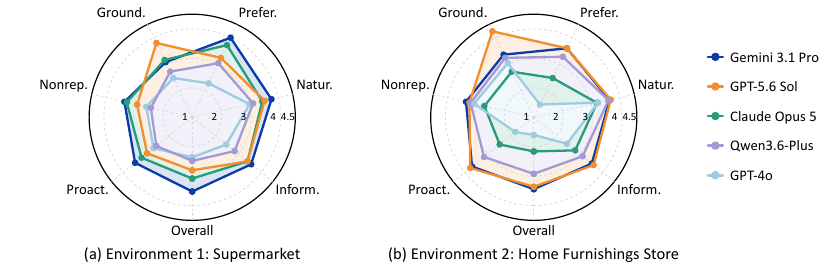}
\caption{Selected dialogue ratings with GPT-4o users (25 tasks per assistant/environment). Both panels display 1--4.5 of the original 1--5 scale.}
\label{fig:ratings_4o}
\end{figure}

\clearpage

\begin{table}[!t]
\centering

\setlength{\tabcolsep}{3pt}
\caption{Purchase quality. GPT-5.6 Sol: four-subset means and SDs for ranking and catalog scores; GPT-4o: means only. RPR uses all episodes; RPR$\mid$Cart conditions on purchase.}
\label{tab:purchase_quality}
{\small
\begin{tabular*}{\linewidth}{@{\extracolsep{\fill}}llcccc@{}}
\toprule
\multirow{2}{*}{\textbf{Environment}} & \multirow{2}{*}{\textbf{Model}} & \multicolumn{2}{c}{\textbf{Outcome}} & \multicolumn{2}{c}{\textbf{Recommendation}} \\
\cmidrule(lr){3-4}\cmidrule(lr){5-6}
 &  & \textbf{RPR (\%)} & \textbf{RPR$\mid$Cart (\%)} & \textbf{R-MRR@10 (\%)} & \textbf{Catalog (\%)} \\
\midrule
\multicolumn{6}{l}{\textbf{GPT-5.6 Sol Users}} \\
\addlinespace[2pt]
\multirow{7}{*}{Supermarket} & Qwen3-VL-8B & 26.0 & 76.5 & 53.7$_{\pm 6.7}$ & 97.9$_{\pm 1.0}$ \\
 & Gemma 4 E4B & 19.0 & 67.9 & 54.7$_{\pm 4.8}$ & 96.4$_{\pm 0.3}$ \\
 & GPT-4o & 42.0 & 93.3 & 70.6$_{\pm 3.6}$ & 99.7$_{\pm 0.2}$ \\
 & Qwen3.6-Plus & 58.0 & 89.2 & 74.6$_{\pm 5.3}$ & 99.4$_{\pm 0.3}$ \\
 & Claude Opus 5 & 62.0 & \textbf{93.9} & \textbf{77.3}$_{\pm 4.9}$ & \textbf{99.9}$_{\pm 0.1}$ \\
 & GPT-5.6 Sol & 64.0 & 92.8 & 77.1$_{\pm 5.3}$ & \textbf{99.9}$_{\pm 0.1}$ \\
 & Gemini 3.1 Pro & \textbf{74.0} & 92.5 & 75.7$_{\pm 3.4}$ & \textbf{99.9}$_{\pm 0.0}$ \\
\midrule
\multirow{7}{*}{\makecell[l]{Home\\Furnishings\\Store}} & Qwen3-VL-8B & 13.0 & 44.8 & 32.0$_{\pm 5.1}$ & 89.4$_{\pm 1.3}$ \\
 & Gemma 4 E4B & 23.0 & 74.2 & 46.6$_{\pm 6.4}$ & 91.6$_{\pm 2.1}$ \\
 & GPT-4o & 20.0 & 57.1 & 40.0$_{\pm 6.5}$ & \textbf{100.0}$_{\pm 0.0}$ \\
 & Claude Opus 5 & 31.0 & 62.0 & 57.4$_{\pm 7.7}$ & \textbf{100.0}$_{\pm 0.1}$ \\
 & Qwen3.6-Plus & 36.0 & 66.7 & 59.3$_{\pm 1.7}$ & \textbf{100.0}$_{\pm 0.0}$ \\
 & GPT-5.6 Sol & 48.0 & 73.8 & \textbf{67.3}$_{\pm 3.0}$ & \textbf{100.0}$_{\pm 0.0}$ \\
 & Gemini 3.1 Pro & \textbf{54.0} & \textbf{81.8} & 64.4$_{\pm 3.8}$ & \textbf{100.0}$_{\pm 0.0}$ \\
\midrule
\multicolumn{6}{l}{\textbf{GPT-4o Users}} \\
\addlinespace[2pt]
\multirow{5}{*}{Supermarket} & GPT-4o & 32.0 & 47.1 & 51.3 & 99.9 \\
 & Qwen3.6-Plus & 40.0 & 76.9 & 64.7 & 99.6 \\
 & GPT-5.6 Sol & 44.0 & 61.1 & 65.4 & \textbf{100.0} \\
 & Claude Opus 5 & 56.0 & 73.7 & 72.2 & 99.8 \\
 & Gemini 3.1 Pro & \textbf{80.0} & \textbf{87.0} & \textbf{73.4} & \textbf{100.0} \\
\midrule
\multirow{5}{*}{\makecell[l]{Home\\Furnishings\\Store}} & GPT-4o & 12.0 & 75.0 & 12.4 & \textbf{100.0} \\
 & Claude Opus 5 & 12.0 & 37.5 & 47.4 & \textbf{100.0} \\
 & Qwen3.6-Plus & 36.0 & 60.0 & 60.1 & \textbf{100.0} \\
 & GPT-5.6 Sol & 52.0 & 72.2 & \textbf{64.0} & \textbf{100.0} \\
 & Gemini 3.1 Pro & \textbf{60.0} & \textbf{78.9} & 62.8 & \textbf{100.0} \\
\bottomrule
\end{tabular*}
}
\end{table}

\subsection{Purchase Quality}\label{app:additional-analysis}
Table~\ref{tab:purchase_quality} evaluates recommendations and purchases against task-specific reference items, including recommendation rankings from interactions without a purchase. Gemini 3.1 Pro most often completes purchases of reference items in both environments with both user simulators. However, with GPT-5.6 Sol users, Claude Opus 5 and GPT-5.6 Sol rank reference items highest in Supermarket and Home Furnishings Store, respectively. Better ranking of suitable products therefore does not always translate into more completed purchases of those products; assistants also need to adapt their recommendations to the preferences users express during the interaction. In Supermarket with GPT-5.6 Sol users, proprietary assistants have similarly high proportions of reference items among completed purchases, suggesting that their main difference lies in how often they carry the interaction through to a purchase.

\definecolor{FailCaseRed}{HTML}{B91C1C}
\definecolor{SuccCaseGreen}{HTML}{047857}
\definecolor{CaseMuted}{HTML}{6B7280}
\newcommand{\failcase}{\textcolor{FailCaseRed}{\textbf{Failure}}}
\newcommand{\succcase}{\textcolor{SuccCaseGreen}{\textbf{Success}}}
\newcommand{\dialogueentry}[2]{\textcolor{CaseMuted}{#1}\quad #2}

\clearpage
\begin{figure}[!t]
\captionsetup{type=figure}
\centering
\includegraphics[width=\linewidth]{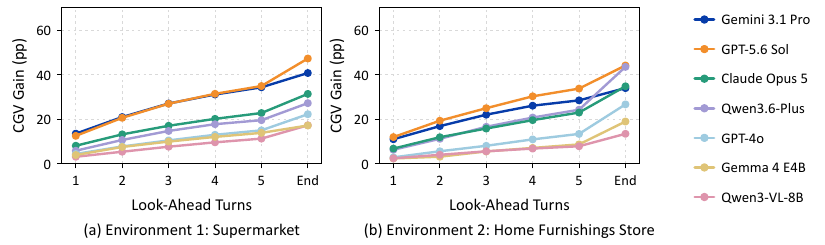}
\caption{Goal visibility gain with GPT-5.6 Sol users: CGV(h) - CGV(0), in percentage points. End extends through the saved episode end.}
\label{fig:cross_environment_cgv}
\end{figure}

\begin{figure}[!t]
\captionsetup{type=figure}
\centering
\includegraphics[width=\linewidth]{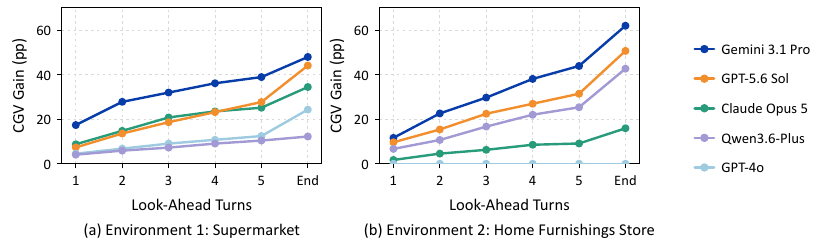}
\caption{Goal visibility gain with GPT-4o users.}
\label{fig:cgv_4o}
\end{figure}

\begin{figure}[!t]
\captionsetup{type=figure}
\centering
\includegraphics[width=\linewidth]{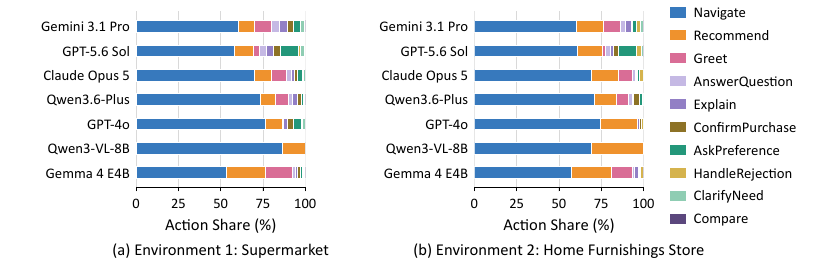}
\caption{Shares of all ten assistant dialogue actions with GPT-5.6 Sol users, pooled over 100 episodes per model and environment. Each response contributes one primary action.}
\label{fig:assistant_behavior}
\end{figure}

\subsection{Navigation and Interaction Behavior}\label{app:failure-analysis}\label{app:error-analysis-details}\label{sec:error_analysis}\label{app:per-action-navigation}
Navigation determines which scene information becomes available in subsequent turns. The differences in CGV gain reflect how well assistants guide users toward making their navigation targets visible. This makes information acquisition an important research direction: choosing when to explore the surroundings and when to approach a visible target so that subsequent recommendations can use the resulting observations (Figures~\ref{fig:cross_environment_cgv} and~\ref{fig:cgv_4o}). CGV is defined in Appendix~\ref{app:metric-definitions}.

Navigate is the most frequent assistant dialogue action across all models (Figure~\ref{fig:assistant_behavior}).

\clearpage
\begin{figure}[!t]
\captionsetup{type=figure}
\centering
\includegraphics[width=\linewidth]{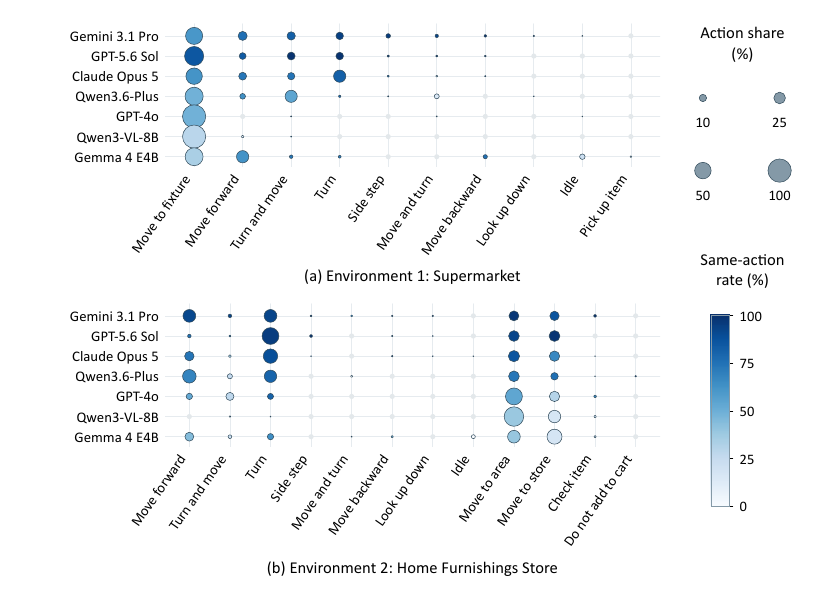}
\caption{Suggested actions within Navigate responses. Bubble area shows action share; color shows the same-action rate. Gray dots indicate zero use.}
\label{fig:navigate_action_following}
\end{figure}

However, assistants affect the environment through users who choose their own actions. In Supermarket, \texttt{move\_to\_fixture} is the most frequently suggested action for every model, but its same-action rate varies substantially (Figure~\ref{fig:navigate_action_following}). GPT-4o and Qwen3-VL-8B suggest this action in almost all of their Navigate responses, yet their corresponding same-action rates are only 48.6\% and 28.7\%. Frequently suggesting an action therefore does not imply that users will execute it.

The differences in same-action rate make adaptation to these choices a concrete research problem. Assistants need to track what the user actually does, interpret the resulting view, and revise their next instruction accordingly. This calls for studying how models maintain effective guidance when user behavior differs from their expectations (Figures~\ref{fig:assistant_behavior} and~\ref{fig:navigate_action_following}).

Reaching the recommendation limit (Rec. Limit) accounts for a substantial share of non-purchase endings, particularly for the two smaller open-weight models (Figure~\ref{fig:termination_reasons}). Together with the cases in Appendix~\ref{app:qualitative-cases}, these results show that repeated recommendation attempts do not ensure a purchase. In the unsuccessful Supermarket case, the assistant recommends a \$10.99 product after the user has requested unbreaded fish within a \$10 budget. In the successful case, inspection and follow-up questions establish that the cod suits the user's cooking needs. Subsequent product choices need to address the requirements users have already stated, including feedback on rejected recommendations.

\clearpage
\begin{figure}[!t]
\captionsetup{type=figure}
\centering
\includegraphics[width=\linewidth]{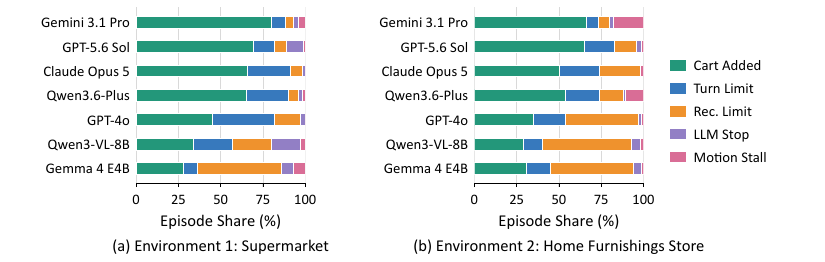}
\caption{Episode termination reasons with GPT-5.6 Sol users (100 tasks per model and environment).}
\label{fig:termination_reasons}
\end{figure}

\begin{table}[!t]
\centering
\setlength{\tabcolsep}{3pt}
\caption{Purchase quality under visual-input ablation. RPR uses all 100 tasks; RPR$\mid$Cart uses purchases.}
\label{tab:ablation_diagnostics}
{\small
\begin{tabular*}{\linewidth}{@{\extracolsep{\fill}}lcc@{}}
\toprule
\textbf{Condition} & \textbf{RPR (\%)} & \textbf{RPR$\mid$Cart (\%)} \\
\midrule
Full Context & \textbf{64.0} & \textbf{92.8} \\
~~ \textit{w/o} EVV & \textbf{64.0} & 90.1 \\
~~ \textit{w/o} EVV \& Textual Caption & 59.0 & 90.8 \\
\bottomrule
\end{tabular*}
}
\end{table}

\subsection{Ablation Protocol}\label{app:ablation-sensitivity}\label{app:visual-ablation-full}
All three conditions use the same 100 profile--goal configurations, with ablations applied to current and historical assistant inputs. In ``w/o EVV,'' we remove the assistant's first-person RGB image but retain textual visible-target annotations. ``w/o EVV \& Textual Caption'' additionally removes visible items, fixtures and navigation surfaces. Both retain the map and pose, product catalog and location index, dialogue, and cart and held-item state. The latter condition adds a static index of all fixture identifiers and supported approach options, without visibility information. Its prompt removes the requirement to use identifiers from the assistant's current view and allows pickup suggestions without product identifiers. Users retain their first-person observations and must select currently actionable targets. Results appear in Table~\ref{tab:visual_ablation}.

Without egocentric images and textual annotations, the assistant gives more directional instructions and fewer instructions targeting specific fixtures. Users may follow these instructions while making limited progress toward the desired products, so higher LNSR does not necessarily correspond to better task outcomes.

Table~\ref{tab:ablation_diagnostics} additionally reports RPR and RPR$\mid$Cart for the three conditions. Removing the egocentric image alone leaves RPR unchanged at 64.0\%, whereas additionally removing textual annotations lowers it to 59.0\%. RPR$\mid$Cart remains above 90\% in all three conditions (90.1--92.8\%). Thus, under the joint ablation, fewer tasks end with a reference-item purchase, even though reference items still account for most completed purchases. Together with Table~\ref{tab:visual_ablation}, these results suggest that retaining first-person information helps more tasks end in a purchase while maintaining a high proportion of reference items among purchases.

\subsection{Qualitative Case Study}\label{app:qualitative-cases}
We examine four GPT-5.6 Sol-user episodes to illustrate how shopping needs shape purchase decisions: mild seafood in Supermarket and the same gray lounge chair for reading or interior design in Home Furnishings Store. The cases involve different tasks and assistants. Tables give their configurations and key dialogue, with views before and after inspection. Turn indices start at zero.
\clearpage
\setbox0=\vbox{\hsize=\linewidth
\begin{minipage}{\linewidth}
\begingroup
\captionsetup{type=table}
\centering
\setlength{\tabcolsep}{3pt}
\caption{Supermarket case overview.}\label{tab:case-supermarket-summary}
{\small
\begin{tabularx}{\linewidth}{@{}p{0.13\linewidth}>{\raggedright\arraybackslash}X>{\raggedright\arraybackslash}X@{}}\toprule
 & \textbf{\failcase} & \textbf{\succcase} \\ \midrule
Task & Task 17.2 & Task 3.3 \\
User & GPT-5.6 Sol & GPT-5.6 Sol \\
Assistant & GPT-4o & Gemini 3.1 Pro \\
Goal & I need another mild dinner protein option. & I need a lighter seafood protein for dinner. \\
Budget & \$10 & \$10 \\
Turn & 8 & 6 \\
Outcome & Empty cart after 8 turns; recommendation attempts exhausted. & Karma Seafood Cod Fillets (\$9.99) added after 6 turns. \\
\bottomrule\end{tabularx}
}
\par\endgroup

\begin{table}[H]
\centering
\setlength{\tabcolsep}{3pt}
\caption{Supermarket dialogue excerpts.}\label{tab:case-supermarket-dialogue}
{\small
\begin{tabularx}{\linewidth}{@{}p{0.13\linewidth}>{\raggedright\arraybackslash}X>{\raggedright\arraybackslash}X@{}}\toprule
\textbf{Stage} & \textbf{\failcase{} Trajectory} & \textbf{\succcase{} Trajectory} \\ \midrule
Initial Need & \dialogueentry{User, T1}{``I'm actually looking for a mild protein that cooks quickly; where should I look?''} & \dialogueentry{User, T1}{``A light seafood dish sounds best.''} \\ \addlinespace[2pt]
Inspection & \dialogueentry{User, T5}{``I'll pass because it's breaded. Do you have a plain, mild fish that cooks quickly for under \$10?''} & \dialogueentry{User, T4}{``Yes, I'd like to check whether the cod is plain and suitable for baking or pan-searing.''} \\ \addlinespace[2pt]
Decision & \dialogueentry{Assistant, T6}{``I recommend the Karma Seafood Tuna, which is plain, mild, and quick to cook. It's priced at \$10.99, which is slightly above your budget but a great option. Would you like to take a closer look at it?''}\par\smallskip \dialogueentry{User, T6}{``That's over my \$10 limit, so I can't take it. Is there any plain, mild fish here that's actually under budget?''} & \dialogueentry{Assistant, T5}{``Yes, these are plain frozen cod fillets, so they would be perfect for baking or pan-searing. Would you like to add them to your cart, or would you prefer to compare them with the tuna or tilapia?''}\par\smallskip \dialogueentry{User, T5}{``The plain cod sounds suitable, so I'll take it.''} \\ \addlinespace[2pt]
Terminal & \dialogueentry{Recorded end}{\texttt{recommendation\_exhausted}} & \dialogueentry{User action}{\texttt{add\_to\_cart}}\newline\dialogueentry{Recorded end}{\texttt{success\_cart\_added}} \\
\bottomrule\end{tabularx}
}
\end{table}

\begin{figure}[H]
\captionsetup{type=figure}
\captionsetup[subfigure]{font=normalsize}
\centering
\begin{subfigure}[t]{0.24\linewidth}\centering\includegraphics[width=\linewidth]{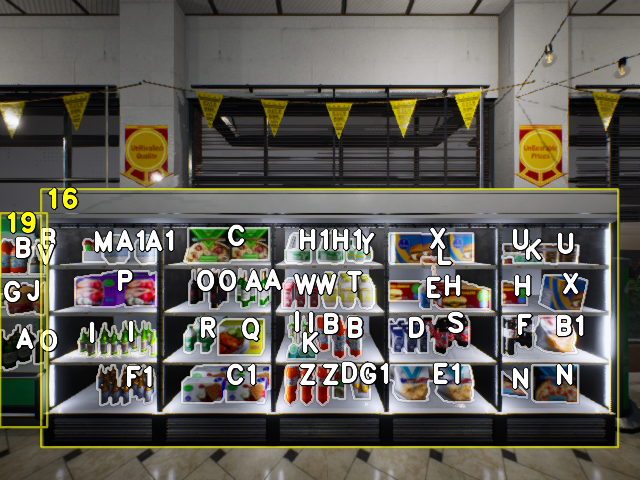}\caption{\failcase{}: T4: fish at the cooler.}\end{subfigure}\hfill
\begin{subfigure}[t]{0.24\linewidth}\centering\includegraphics[width=\linewidth]{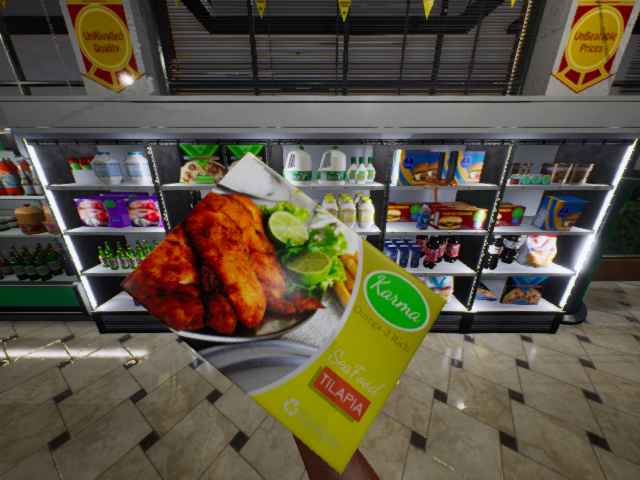}\caption{\failcase{}: T5: holding breaded tilapia.}\end{subfigure}\hfill
\begin{subfigure}[t]{0.24\linewidth}\centering\includegraphics[width=\linewidth]{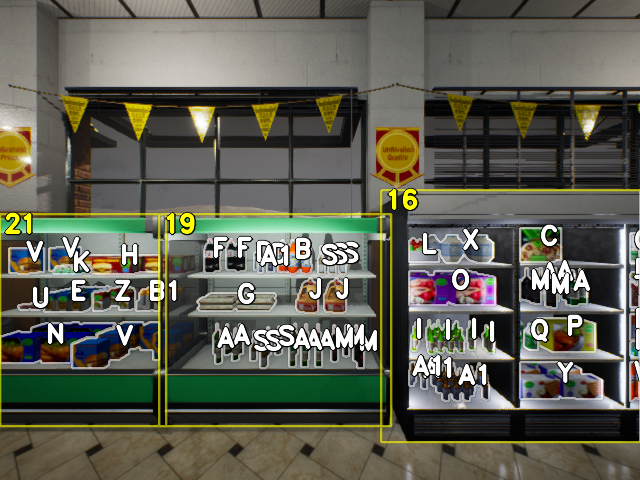}\caption{\succcase{}: T4: fish at the cooler.}\end{subfigure}\hfill
\begin{subfigure}[t]{0.24\linewidth}\centering\includegraphics[width=\linewidth]{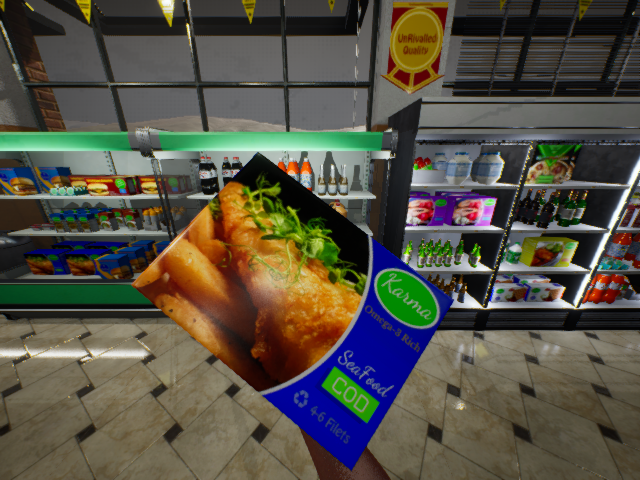}\caption{\succcase{}: T5: holding cod fillets.}\end{subfigure}\hfill
\caption{Seafood inspection: breaded tilapia is rejected; plain cod is purchased.}\label{fig:case-supermarket}
\end{figure}

The Supermarket cases show how user feedback needs to inform subsequent product choices. In the unsuccessful episode, the assistant recommends an over-budget product after the user specifies a preference for unbreaded fish and a budget limit. In the successful episode, inspection and follow-up questions establish that the cod suits the user's cooking needs and help complete the purchase.
\end{minipage}}
\typeout{CASEHEIGHT-SUPERMARKET: \the\ht0; DEPTH: \the\dp0; LIMIT: \the\textheight}
\box0

\clearpage
\setbox0=\vbox{\hsize=\linewidth
\begin{minipage}{\linewidth}
\begingroup
\captionsetup{type=table}
\centering
\setlength{\tabcolsep}{3pt}
\caption{Home Furnishings Store case overview.}\label{tab:case-furnishings-summary}
{\small
\begin{tabularx}{\linewidth}{@{}p{0.13\linewidth}>{\raggedright\arraybackslash}X>{\raggedright\arraybackslash}X@{}}\toprule
 & \textbf{\failcase} & \textbf{\succcase} \\ \midrule
Task & Task 24.3 & Task 2.4 \\
User & GPT-5.6 Sol & GPT-5.6 Sol \\
Assistant & Claude Opus 5 & GPT-5.6 Sol \\
Goal & Maybe a warmer rounded seat would be better. & I need somewhere comfortable to read. \\
Budget & \$310 & \$310 \\
Turn & 8 & 4 \\
Outcome & Empty cart after 8 turns; recommendation attempts exhausted. & Charcoal-Piped Gray Wood-Frame Lounge Chair (\$227) added after 4 turns. \\
\bottomrule\end{tabularx}
}
\par\endgroup

\begin{table}[H]
\centering
\setlength{\tabcolsep}{3pt}
\caption{Home Furnishings Store dialogue excerpts.}\label{tab:case-furnishings-dialogue}
{\small
\begin{tabularx}{\linewidth}{@{}p{0.13\linewidth}>{\raggedright\arraybackslash}X>{\raggedright\arraybackslash}X@{}}\toprule
\textbf{Stage} & \textbf{\failcase{} Trajectory} & \textbf{\succcase{} Trajectory} \\ \midrule
Initial Need & \dialogueentry{User, T0}{``I'm looking for a chair to make our apartment feel more comfortable.''} & \dialogueentry{User, T0}{``I'm looking for comfortable seating for reading, ideally something I can sit in for a while.''} \\ \addlinespace[2pt]
Inspection & \dialogueentry{User, T3}{``This chair looks promising for the apartment; I'd like to see whether its comfort and upholstery feel right.''} & \dialogueentry{User, T2}{``This looks like the kind of chair I could settle into with a book. I'd like to confirm its comfort, size, and price before deciding.''} \\ \addlinespace[2pt]
Decision & \dialogueentry{Assistant, T4}{``This one checks out nicely — it's a Scandinavian midcentury lounge chair with a natural wood frame and gray fabric upholstery with charcoal piping, priced at \$227. It's sized for a single seat and works well in a living room or bedroom. I'd go ahead and add it to your cart.''}\par\smallskip \dialogueentry{User, T4}{``The chair seems comfortable, but the cool gray and sharper shape aren't quite right for warming up our apartment, so I'll skip it.''} & \dialogueentry{Assistant, T3}{``This chair is a strong fit for extended reading: it’s a single lounge chair with gray fabric upholstery, natural wood open arms, and a splayed-leg frame. The confirmed price is 227 USD, so I recommend adding it to your cart.''}\par\smallskip \dialogueentry{User, T3}{``This chair sounds comfortable enough for longer reading, and at \$227 it fits my budget. I'll take it.''} \\ \addlinespace[2pt]
Terminal & \dialogueentry{Recorded end}{\texttt{recommendation\_exhausted}} & \dialogueentry{User action}{\texttt{add\_to\_cart}}\newline\dialogueentry{Recorded end}{\texttt{success\_cart\_added}} \\
\bottomrule\end{tabularx}
}
\end{table}

\begin{figure}[H]
\captionsetup{type=figure}
\captionsetup[subfigure]{font=normalsize}
\centering
\begin{subfigure}[t]{0.24\linewidth}\centering\includegraphics[width=\linewidth]{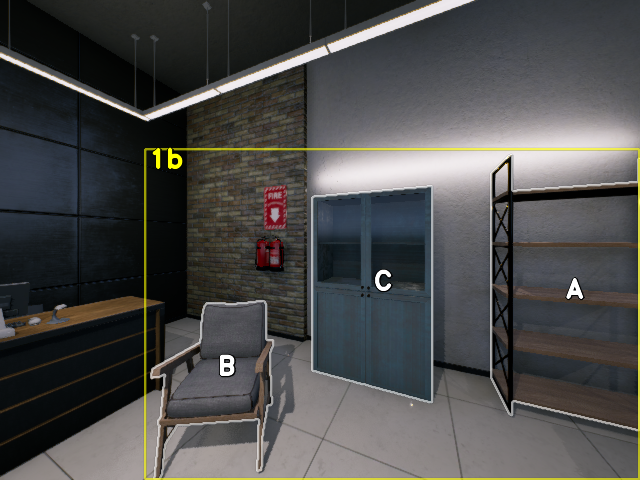}\caption{\failcase{}: T3: gray chair in view.}\end{subfigure}\hfill
\begin{subfigure}[t]{0.24\linewidth}\centering\includegraphics[width=\linewidth]{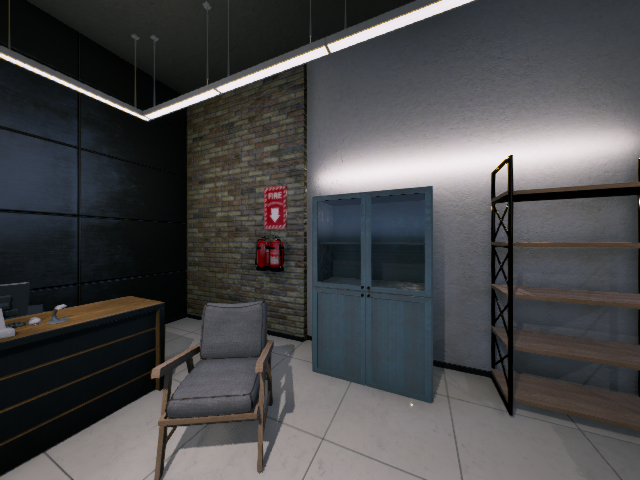}\caption{\failcase{}: T4: after checking the gray chair.}\end{subfigure}\hfill
\begin{subfigure}[t]{0.24\linewidth}\centering\includegraphics[width=\linewidth]{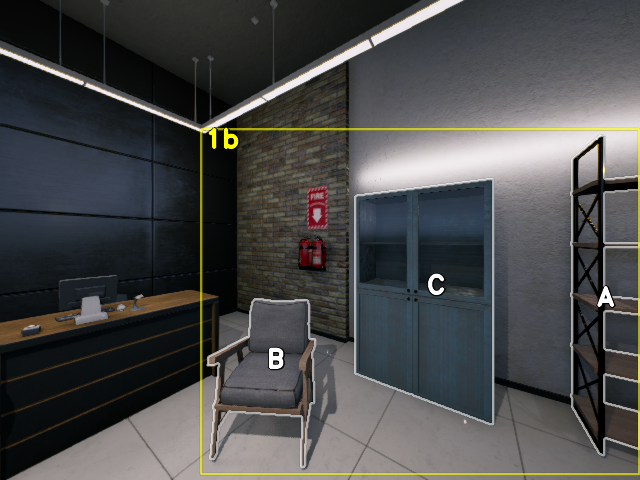}\caption{\succcase{}: T2: gray chair in view.}\end{subfigure}\hfill
\begin{subfigure}[t]{0.24\linewidth}\centering\includegraphics[width=\linewidth]{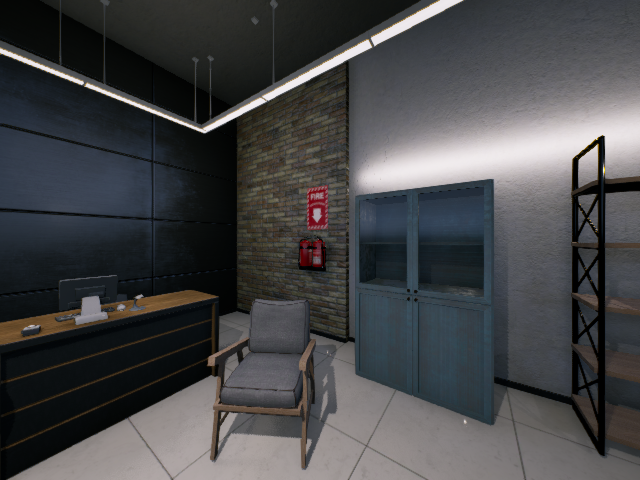}\caption{\succcase{}: T3: after checking the gray chair.}\end{subfigure}\hfill
\caption{Chair inspection: gray upholstery is rejected for a warm interior and accepted for a reading corner.}\label{fig:case-furnishings}
\end{figure}

The Home Furnishings Store cases illustrate how intended use shapes product choices. The same gray lounge chair is accepted for reading because of its perceived comfort and price, but rejected for a warmer interior because of its color and shape. Effective recommendations connect product features to the experience the user wants.
\end{minipage}}
\typeout{CASEHEIGHT-FURNITURE: \the\ht0; DEPTH: \the\dp0; LIMIT: \the\textheight}
\box0

\clearpage
\FloatBarrier
\section{Prompt Templates}
\label{app:prompt-templates}

The following templates summarize the runtime instructions and output schemas for the recommendation assistant, user simulator, dialogue moderator, and conversation judge. The released benchmark includes the complete assembled prompts.

\assistantprompttemplate
\userprompttemplate
\moderatorprompttemplate
\judgeprompttemplate
\FloatBarrier